\documentclass{article}

 \usepackage[main, final]{neurips_2026}
\usepackage[utf8]{inputenc} % allow utf-8 input
\usepackage[T1]{fontenc}    % use 8-bit T1 fonts
\usepackage{amsfonts}       % blackboard math symbols
\usepackage{array}
\usepackage{booktabs}       % professional-quality tables
\usepackage{caption}
\usepackage{enumitem}
\usepackage{setspace}
\usepackage{float}
\usepackage{fontawesome5}

\usepackage{graphicx}
\usepackage{microtype}      % microtypography
\usepackage{multirow}
\usepackage{nicefrac}       % compact symbols for 1/2, etc.
\usepackage{placeins}
\usepackage{siunitx}
\usepackage{graphicx}
\usepackage[ruled,vlined]{algorithm2e}
\usepackage{epstopdf}
\usepackage{amsmath}
\usepackage{tikz}
\usepackage{subcaption}
\usepackage{wasysym}
\usepackage[table]{xcolor} 
\usepackage{url}            % simple URL typesetting
\usepackage{hyperref}       % hyperlinks
\SetKwInput{KwInput}{Input}
\SetKwInput{KwOutput}{Output}
\usepackage[table]{xcolor}
\SetKwIF{If}{ElseIf}{Else}{if}{then}{else if}{else}{end}
\usepackage{array}
\newcolumntype{C}[1]{>{\centering\arraybackslash}m{#1}}
\title{UFO: A Unified Flow-Oriented Framework for Robust Continual Graph Learning}
\author{Danhui Zhang$^{1}$, Zhe Wang$^{1}$, Qing Qing$^{1}$, Jiarui Liu$^{1}$, Wentao Gao$^{2}$, Ziqi Xu$^{3}$,\\
\textbf{Mingliang Hou}$^{4}$, \textbf{Xikun Zhang}$^{3}$, \textbf{Renqiang Luo}$^{1}$\\
$^1$Jilin University, $^2$Adelaide University, $^3$RMIT University, $^4$Jinan University
}
\begin{document}

\maketitle

\begin{abstract}
Graph learning research has increasingly shifted toward continual graph learning (CGL), which better reflects real-world scenarios where graphs evolve over time. However, existing CGL methods largely assume clean supervision and overlook a critical challenge: the newly arriving portions of the graph are often noisy, due to annotation errors or adversarial corruption. This mismatch limits their applicability in practice.
In this work, we study robust continual graph learning, where models must simultaneously handle catastrophic forgetting and noisy supervision in evolving graph data. We show that label noise introduces a new failure mode—catastrophic remembering, where models persistently reinforce corrupted knowledge across tasks.
To address these challenges, we propose a Unified Flow-Oriented framework (UFO). First, UFO models conditional feature distributions via flow-based generative modeling and produces replay representations, mitigating forgetting without storing historical data. Second, UFO estimates instance-level reliability scores to distinguish clean from noisy nodes, reducing the impact of corrupted supervision and alleviating catastrophic remembering.
Extensive experiments on four benchmark graph datasets under varying noise ratios demonstrate that UFO consistently outperforms existing methods in both accuracy and forgetting metrics. Code is available at: \url{https://anonymous.4open.science/r/UFO}.

\iffalse
Label noise is pervasive in the real world, making continual learning fundamentally different. 
It requires the model to identify reliable information over a sequence of tasks. 
Existing methods are designed to balance stability and plasticity, but achieving this objective depends on high quality training data. 
Due to human mistakes and malicious attacks, such an assumption severely limits the application in practice.
In this paper, we consider a more practical and challenging scenario of robust continual graph learning. 
Specifically, catastrophic forgetting commonly occurs in continual learning,
while label noise introduces an additional disruption to the model, which results in catastrophic remembering.
To address these challenges, we design a \textbf{U}nified \textbf{F}low-\textbf{O}riented framework (\textbf{UFO}) in two aspects:
(1) flow models conditional feature distributions and generates replay representations, mitigating catastrophic forgetting without storing historical examples;
(2) flow computes instance level scores for clean and noisy nodes,
mitigating the effect of corrupted supervision and catastrophic remembering.
We support our design with extensive experiments on four graph datasets under different noise ratios.
The results show that UFO outperforms existing methods in terms of both accuracy and forgetting.
Our code is available at: \url{https://anonymous.4open.science/r/UFO}.
\fi
\end{abstract}

% For example, in product networks, new products are introduced and new relations are formed, continuously adding nodes and edges to the graph. 

\section{Introduction}
\label{sec:introduction}
\par Graphs in real-world applications are inherently dynamic and evolve over time~\cite{xia2026graph}. 
In social, citation, and co-authorship networks, new entities and connections emerge continuously \cite{bojchevski2018deep}. 
To handle this data stream, models must acquire new knowledge while preserving previously learned information. 
This necessity has established Continual Graph Learning (CGL) as a critical research area \cite{kirkpatrick2017overcoming}, focusing on learning from sequential graph data.

\par Despite its potential, CGL faces two primary obstacles. 
First, storage limitations and privacy concerns often make historical data inaccessible, leading to catastrophic forgetting \cite{li2024matters}. 
Second, while most CGL models assume clean training labels, label noise is inevitable in practice. 
Real-world datasets often contain significant noise ratios, sometimes exceeding $30$\% \cite{song2019selfie}. 
On graphs, this noise is particularly destructive as the message-passing mechanism propagates and amplifies errors across neighborhoods \cite{wang2024noisygl}. 
This forces the model to memorize incorrect information, causing catastrophic remembering \cite{xu2024mitigate}, 
where noise from new tasks disrupts the knowledge learning.
% where noise from new tasks conflicts with old knowledge.

\par Current techniques rarely address both challenges simultaneously.
% ~\cite{li2017learning}. 
Standard CGL methods rely on regularization or replay buffers to prevent forgetting \cite{zhang2023hierarchical},
yet they neglect the quality of labels and fail to handle noise.
% yet they are overconfident in the accuracy of new labels. 
Conversely, robust learning methods use sample selection or robust loss functions to mitigate noise \cite{han2018co}, but they typically focus on the current task and ignore historical knowledge.
Therefore, their effectiveness is strongly tied to accessible data, leaving knowledge learned from inaccessible data vulnerable to forgetting~\cite{qi2024revealing}.
While a few hybrid approaches \cite{xu2024mitigate,cao2025erroreraser,kim2021continual} have been proposed for image data, 
they cannot perform as well on graphs.
This is because noise propagates through the message-passing mechanism.
The complex topology of graphs further complicates the distinction between noise and legitimate structural shifts~\cite{li2025leveraging}.

\par To bridge this gap, we propose a unified perspective in which catastrophic forgetting and noise robustness can be addressed simultaneously. 
Our key insight is that both challenges can be mitigated by learning and leveraging the underlying feature distributions of the data. 
By aligning new observations with these distributions, the model can suppress noisy supervision, while the distributions themselves enable knowledge retention without storing historical data.

Concretely, we replace explicit data storage with a generative distribution model. 
This model synthesizes representative features of past tasks for replay, preserving performance while avoiding the need to retain sensitive or large-scale historical data. 
At the same time, we estimate the reliability of incoming nodes based on their consistency with the learned distributions, allowing the model to down-weight noisy or corrupted samples during training.

Building on this idea, we propose \textbf{UFO}, a unified framework for robust continual graph learning without historical data storage. 
UFO models conditional feature distributions to generate replay representations, effectively mitigating catastrophic forgetting. 
To handle label noise, UFO incorporates an instance-level adaptation mechanism that assigns importance weights to nodes based on their alignment with the learned feature space, ensuring that reliable samples dominate model updates. 
In addition, a knowledge preservation strategy constrains representation drift across tasks, maintaining both structural and semantic consistency over time.

In summary, our main contributions are as follows:
\begin{itemize}[leftmargin=0.5cm]
    \item We introduce the problem of robust continual graph learning with label noise, and identify a new failure mode, catastrophic remembering, caused by the accumulation of corrupted supervision.
    \item We propose UFO, a unified framework that combines distribution-based feature replay and instance-level reliability estimation to address forgetting and noisy labels without storing historical data.
    \item We conduct extensive experiments on multiple benchmark datasets, demonstrating that UFO consistently improves both predictive performance and robustness under varying noise levels.
\end{itemize}

\section{Related Work}
\subsection{Continual Graph Learning}
Continual graph learning \cite{zhang2024continual} is a paradigm that enables models to learn from evolving graphs while mitigating catastrophic forgetting.
Existing approaches are divided into three families:
(1) Regularization-based methods introduce a penalty term in the loss function to constrain the updates of crucial parameters.
Early studies like LwF \cite{li2017learning} learn the soft labels from the old model and TWP \cite{liu2021overcoming} preserves the topological aggregation of previous graphs.
(2) Experience replay-based methods store or generate representative samples from previous tasks,
e.g., DMSG \cite{qiao2025towards} takes into account memory diversity through selection and generation.
DSLR \cite{choi2024dslr} introduces the coverage-based diversity and graph structure learning. 
(3) Parameter isolation-based methods use different parameters to different tasks, so that learning a new task has less effect on old tasks.
HPNs \cite{zhang2023hierarchical} dynamically learns different levels of prototypes.
However, all these methods are dependent on clean supervision,
which limits their practicality in real-world scenarios \cite{sheng2025ca2c}.
In this work, we consider a more challenging problem in robust continual graph learning: 
how to learn from sequential tasks under corrupted supervision when historical data are inaccessible?
This requires the model to preserve correct knowledge from seen tasks while avoiding the reinforcement of corrupted information.

\subsection{Graph Learning with Label Noise}
Learning with label noise \cite{song2023learning} focuses on training models with corrupted data. 
There have been many efforts to enhance robustness, such as robust loss design~\cite{zhang2018generalized}, label correction \cite{yi2019probabilistic}, and sample selection \cite{yu2019how}.
For example, SCE \cite{wang2019symmetric} improves robustness by combining cross entropy with a
reverse cross entropy.
Bootstrap \cite{reed2014training} replaces the target labels with a combination of their predicted labels.
Recent works have extended these ideas to graph learning.
CLNode \cite{wei2023clnode} proposes a curriculum learning framework for node classification with a multi-perspective difficulty measurer.
TFR \cite{wang2025learning} leverages a backbone GNN and a decoder GNN to reconstruct topological features to maximize mutual information.
TSS \cite{wu2024mitigating} utilizes graph topology to progressively select confident nodes and trains the model from easy to hard.
However, existing methods focus on the current task and overlook real-world dynamic graph environments, where historical graph data may be unavailable due to privacy or storage constraints,
making previously learned knowledge vulnerable to forgetting.
% The idea of robust loss methods is to redesign the training objective.
% SCE \cite{wang2019symmetric} combines with a reverse cross-entropy term and ELR \cite{liu2020early} introduces a regularization term.
% Label correction methods aim to identify and correct the noisy labels before training.
% Sample selection methods filter out noisy samples using some strategies.
% For instance, 
% Co-teaching \cite{han2018coteaching} trains two networks simultaneously and updates the parameters with exchanged samples.
% designed for static graph learning, where historical data are available in a single stage. 

\section{Problem Statement}
\subsection{Notations}
Consider a sequence of $n$ tasks $\{\mathcal{T}_1, \mathcal{T}_2, \cdots,\mathcal{T}_t,\cdots, \mathcal{T}_n\}$. 
Each task $\mathcal{T}_t$ is defined on a graph $\mathcal{G}_t = \{\mathcal{V}_t, \mathcal{E}_t\}$, where $\mathcal{V}_t$ is the node set and $\mathcal{E}_t$ is the edge set.
The graph structure is represented by the adjacency matrix $A_t \in \mathbb{R}^{N_t \times N_t}$, where $N_t = |\mathcal{V}_t|$. 
Each node $v_i \in \mathcal{V}_t$ has an input feature $h_i^{(0)}$ and an observed label $y_i$.
% We denote the feature matrix as $X_t \in \mathbb{R}^{N_t \times d}$. 
During the training stage of task $\mathcal{T}_t$, the node set $\mathcal{V}_t$ is divided into a training set $\mathcal{V}_t^{tr}$, a validation set $\mathcal{V}_t^{val}$, and a test set $\mathcal{V}_t^{te}$. 
The observed labels may be noisy in $\mathcal{V}_t^{tr}$ and clean in $\mathcal{V}_t^{val}$ and $\mathcal{V}_t^{te}$. 
Formally, our objective is to learn sequential tasks under corrupted supervision while preserving previously learned knowledge:
$\min\limits_{\Theta_t}
\frac{1}{t}\sum_{j=1}^{t}\mathcal{L}_j(\Theta_t)$,
where $\Theta_t$ refers to the model parameters for the $t$-th task.
% UFO consists of a flow $f$, a feature extractor $H$ and a classifier $C$. 
% Our goal is to train a robust continual graph learning model that learns sequential tasks under corrupted supervision  while preservings previously learned knowledge:
\subsection{Preliminary}
Given an input $x\in\mathcal{X}$ and a latent variable $z\in\mathcal{Z}$, a normalizing flow \cite{rezende2015variational,dinh2017density} defines an invertible mapping  $f:\mathcal{X}\rightarrow\mathcal{Z}$ that transforms the original distribution into a base distribution, such as a standard Gaussian distribution.
Here, $z=f(x)$ and $x=f^{-1}(z)$.
By the change of variables formula,
the likelihood of $x$ can be written as:
\begin{equation}\label{eq1}
p_X(x) = p_Z(z) \left| \det \frac{\partial f(x)}{\partial x}\right|,
\end{equation}
where $\left| \det \frac{\partial f(x)}{\partial x}\right|$  is the absolute Jacobian determinant of the transformation $f$. 
In addition, let $y\in\mathcal{Y}$ be the class label,
and the flow can be extended to the conditional setting~\cite{izmailov2020semi}.
The invertible mapping becomes $f:\mathcal{X}\times\mathcal{Y}\rightarrow\mathcal{Z}$, where $z=f(x,y)$ and $x=f^{-1}(z,y)$.
We can express the conditional density of $x$ by:
\begin{equation}\label{eq2}
p_{X\mid Y}(x\mid y) = p_Z(z) \left| \det \frac{\partial f(x,y)}{\partial x}\right|.
\end{equation}

In general, a single transformation is not flexible enough to model a complex distribution \cite{papamakarios2021normalizing}.
Therefore, $f$ is defined as a composition of $K$ simple invertible transformations: $f=f_K \circ \cdots \circ f_1$.
The $\circ$ denotes function composition. 
Based on Eq.~(\ref{eq2}), the conditional log-likelihood of $x$ under label $y$ is:
\begin{equation}
  \log p_{X\mid Y}(x\mid y)=
  \log p_Z(z)+\sum_{k=1}^{K}\log \left|\det \frac{\partial z_k}{\partial z_{k-1}}\right|,
\end{equation}
where $z_k$ is the intermediate variable after the $k$-th transformation.

\section{Methodology}
\label{sec:method}
\begin{figure}
    \centering
    \includegraphics[width=0.95\linewidth]{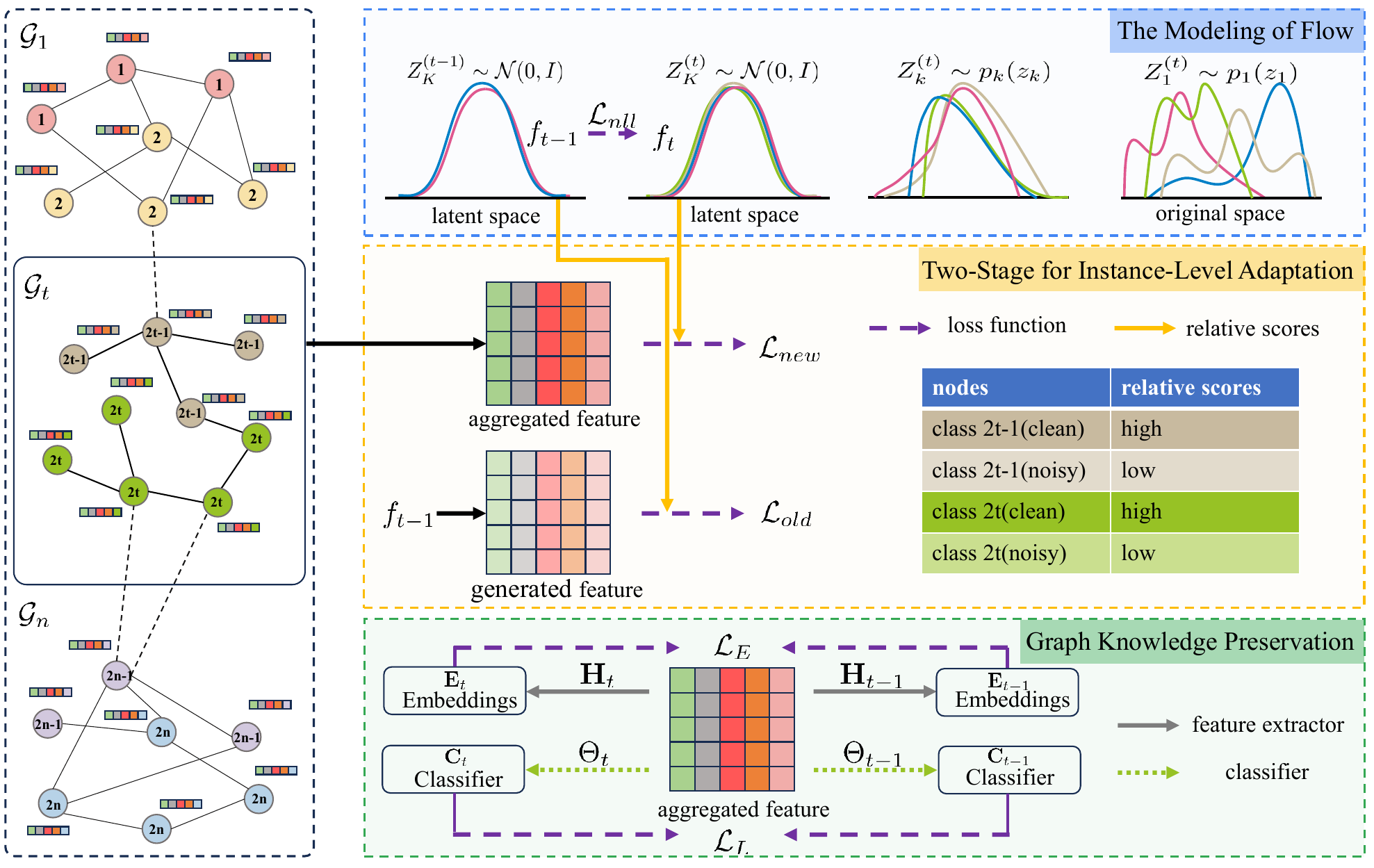}
    \caption{
    The framework of UFO:
    (1) The flow models feature distributions over sequential graphs and generates features for previous tasks.
    (2) Relative scores are computed using $f_t$ and $f_{t-1}$, and are used to modulate $\mathcal{L}_{\mathrm{new}}$ and $\mathcal{L}_{\mathrm{old}}$ for instance-level adaptation.
    (3) Knowledge preservation constraints maintain structural and semantic consistency across tasks.
    }
    \label{fig:placeholder}
\end{figure}

This section introduces UFO, as illustrated in Figure~\ref{fig:placeholder}.
The following sections describe three related modules that work together.
First, Section~\ref{sec:flow_modeling} introduces the modeling of flow, which maps feature distributions into a shared standard Gaussian latent space and uses the learned distributions to generate replay representations for earlier tasks.
Second, Section~\ref{sec:Two-Stage-for} presents the instance-level adaptation mechanism.
At task $\mathcal{T}_t$, the current flow $f_t$ assigns dynamic scores to instances according to their alignment with the task distribution, reducing the effect of noisy supervision in $\mathcal{L}_{\mathrm{new}}$. 
The frozen flow $f_{t-1}$ scores generated replay features in the historical feature space, down-weighting low-likelihood samples in $\mathcal{L}_{\mathrm{old}}$.
Finally, Section~\ref{sec:Graph-Knowledge-Preservation} introduces graph knowledge preservation, which constrains structural and semantic consistency across tasks.

\subsection{The Modeling of Flow }
\label{sec:flow_modeling}
Flows are a type of powerful generative models. 
They learn an invertible mapping from a target data distribution to a simple latent distribution \cite{dinh2017density}. 
Based on this property, we can replay graph features instead of storing exemplars,
mitigating catastrophic forgetting while preserving data privacy to some extent.
Since graph features from different classes follow distinct distributions \cite{gu2025neubm}, 
we employ a conditional normalizing flow to model feature distributions for feature replay.

\par To learn informative graph representations, we employ the graph neural network proposed by \cite{hamilton2017inductive} as the graph encoder. 
Given the neighborhood set $\mathcal{N}(v)$ of node $v$, the embedding $h_v^{(l)}$ at the $l$-th layer is calculated as follows:
\begin{equation}
h_{\mathcal{N}(v)}^{(l)}=\mathrm{AGGREGATE}^{(l)}\bigl( \bigl\{ h_u^{(l-1)},\ \forall u \in \mathcal{N}(v) \bigr\} \bigr),
\end{equation}
\begin{equation}\label{eq5}
    h_v^{(l)}=\sigma\left(W^{(l)} \cdot \mathrm{CONCAT}\bigl(h_v^{(l-1)}, h_{\mathcal{N}(v)}^{(l)}\bigl)\right),
\end{equation}
where $W^{(l)}$ is the learnable parameter at the $l$-th layer.
$\sigma$ is a nonlinear activation function.
Eq.~(\ref{eq5}) shows that the embedding of node $v$ is updated by combining its own representation with the aggregated information from its neighbors.
Following this message-passing mechanism, we construct the aggregated feature $x=\mathrm{CONCAT}\bigl(h_v^{(l-1)}, h_{\mathcal{N}(v)}^{(l)}\bigl)$.

\par When the observed labels are corrupted, the supervision signal becomes unreliable, as label noise weakens the true correlation between features and labels~\cite{cheng2021learning}.
To capture the consistency, the flow models the conditional distribution $p_{X\mid Y}(x\mid y)$ for each class.
The resulting likelihood is used to estimate the reliability of incoming nodes, while the learned distribution synthesizes replay features of past tasks without historical data storage.

For task $\mathcal{T}_t$, the flow adopts a shared standard Gaussian prior $\mathcal{N}(0,I)$ in the latent space and uses $y$ as conditional information.
It is trained on the current task $\mathcal{T}_t$ by minimizing the negative log-likelihood:
\begin{equation}
    \mathcal{L}_{nll}
    =
    -\sum_{v_i\in \mathcal{V}_t^{tr}}
    \log p_{X\mid Y}(x_i \mid y_i).
\end{equation}
However, directly training the flow model on a single task leads to distributional bias. 
Inspired by \cite{wuerkaixi2024accurate}, we incorporate a replay strategy to mitigate this issue. 
In particular, for each previous task $\mathcal{T}_j$ with $1 \le j < t$,
latent variables $\tilde{z}_i$ are sampled from the latent prior conditioned on labels $\tilde{y}_i$,
resulting in the synthetic batch
$\tilde{B}_j=\{(\tilde{z}_i,\tilde{y}_i)\}_{i=1}^{b}$,
where $b$ is the batch size.
To avoid interference from the current task, we use the frozen previous flow to generate features $f_{t-1}^{-1}(\tilde{z}_i,\tilde{y}_i)$.
Note that we use a tilde over the corresponding symbol to denote synthetic variables.
The flow is then optimized using both aggregated and generated features:
% Then the corresponding generated features are used to optimize the flow model with the following objective:
\begin{equation}
   \mathcal{L}_{nll}
    =-\sum_{v_i\in \mathcal{V}_t^{tr}}
    \log p_{X\mid Y}(x_i \mid y_i)-\!\sum_{(\tilde{z}_i,\tilde{y}_i)\in \tilde{B}_j}
    \log p_{X\mid Y}(f_{t-1}^{-1}(\tilde{z}_i, \tilde{y}_i) \mid \tilde{y}_i).
\end{equation}

\subsection{Two-Stage for Instance-Level Adaptation}
\label{sec:Two-Stage-for}
In continual learning with label noise, corrupted labels can degrade the performance of graph neural networks and disrupt the model's inherent stability, as noisy information may spread through graph structure \cite{dai2021nrgnn,zhang2025handling}.
This effect can accumulate across tasks and cause catastrophic remembering, where the model retains incorrect knowledge from corrupted supervision.
To this end, in the first stage, we calculate the conditional log-likelihood
$r_i = \log p_{X \mid Y}(x_i \mid y_i)$ for instance-level adaptation.
% and compute the maximum score $m(\mathbf{r})$. 
Since softmax is shift-invariant:
$\operatorname{softmax}(\mathbf{r})=
\operatorname{softmax}(\mathbf{r} - m(\mathbf{r}))$, the relative scores $\mathbf{s}$ are computed as:
\begin{equation}
  m(\mathbf{r}) := \max_i r_i,\qquad
  \mathbf{s} =b\cdot \operatorname{softmax}\left(\mathbf{r} - m(\mathbf{r})\right),
\end{equation}
where $b$ is the batch size of the current task.
Additionally, when learning a new task,
the scores are smoothed to reduce outliers during early flow training and clipped to the interval $[\alpha,\beta]$: 
$ s_i = \min\left(\beta, \max\left(\alpha, s_i\right)\right)$. 

In the next stage, we optimize two objectives.
First, to fit the new task $\mathcal{T}_t$ under noisy supervision,
we reduce the negative impact of label noise using dynamic scores.
The current flow \(f_t\) evaluates the conditional log-likelihood \(r_i\) to measure the alignment with the current task distribution, 
and \(r_i\) is further converted into the relative score \(s_i\).
The objective is formulated as follows:
\begin{equation}
\mathcal{L}_{ce}^{new} = 
\sum_{v_i \in \mathcal{V}_t^{tr}} s_i\ell_{ce}\left(C_t(H_t(v_i)), y_i\right),
\end{equation}
where $H_t$ is the feature extractor and $C_t$ is the classifier.
$\ell_{ce}(\cdot)$ denotes the cross-entropy loss.
For previous task $\mathcal{T}_j$ with $1 \le j < t$, the second objective uses
the synthetic batch $\tilde{B}_j$ introduced above.
The score $\tilde{s}_i$, computed by the frozen flow $f_{t-1}$, reflects the
alignment of a generated feature with the learned historical feature space,
reducing the influence of low-likelihood or out-of-distribution replay features
caused by imperfect flow generation.
The generative replay loss is defined as:
\begin{equation}
\mathcal{L}_{ce}^{old} = 
\sum_{(\tilde{z}_i, \tilde{y}_i) \in \tilde{B}_j}
\tilde{s}_i \ell_{ce}\left(
C_t\left(f_{t-1}^{-1}(\tilde{z}_i, \tilde{y}_i)\right), 
\tilde{y}_i
\right).
\end{equation}

\begin{algorithm}[t]
  \setstretch{1.15}
  \SetAlgoLined
  \caption{Framework of UFO}
  \label{alg:flow_replay}
  \KwInput{A sequence of $n$ graphs $\{\mathcal{G}_1,\mathcal{G}_2,\dots,\mathcal{G}_n\}$, flow $f$, feature extractor $H$, classifier $C$}
  \KwOutput{Updated $H_n$ and $C_n$}

    \For{$t \gets 1$ \KwTo $n$}{
        \uIf{$t = 1$}{

            Aggregate features $\mathbf{x}_1$ from $\mathcal{G}_1$\;
            Train $f_1$ on $\mathbf{x}_1$ in Eq.~(6)\;
            Train $H_1$ and $C_1$ on $\mathcal{G}_1$ in Eq.~(9)\;

        }
        \Else{
            Aggregate features $\mathbf{x}_t$ from $\mathcal{G}_t$\;
            Generate features $\tilde{\mathbf{x}}_{t-1}$ from $f_{t-1}$\;
            
            Train $f_t$ on $\mathbf{x}_t$ and $\tilde{\mathbf{x}}_{t-1}$ in Eq.~(7)\;
            
            Compute the relative scores from $f_t$ and $f_{t-1}$ in Eq.~(8)\;
            
            Train $H_t$ and $C_t$ on $\mathcal{G}_t$ and $\tilde{\mathbf{x}}_{t-1}$ in Eqs.~(9),~(10),~(13)\;
        }
    }
\end{algorithm}

\subsection{Graph Knowledge Preservation}
\label{sec:Graph-Knowledge-Preservation}
As a sequence of graphs arrives, the original feature space gradually shifts, 
which alters the representations learned from previous tasks and weakens the stability of the latent space \cite{li2017learning}.
Since the flow relies on a stable feature space to model distributions, we preserve graph knowledge to control the shifts and maintain the learned structural and semantic information \cite{yang2020distilling}.
The embeddings are used to anchor the representation of the same node $v_i$:
\begin{equation}
    \mathcal{L}_{E}=\sum_{v_i \in \mathcal{V}_t^{tr}} \left\| H_t(v_i)- H_{t-1}(v_i) \right\|_2^2,
\end{equation}
where  $H_{t-1}$ is the feature extractor from task $t-1$.
Class semantic knowledge is encoded in the output distribution of the classifier, 
which reflects the relations among classes.
Directly optimizing the current classifier may disrupt this previously learned knowledge.
We therefore use the soft predictions from the frozen classifier $C_{t-1}$ to guide the current classifier $C_t$. 
The two classifiers are evaluated with the same task head, and the Kullback--Leibler (KL) divergence is used for distillation:
\begin{equation}
\mathcal{L}_{L}
=
\sum_{v_i \in \mathcal{V}_t^{tr}}
\tau^2
\mathrm{KL}\left(
\operatorname{softmax}(C_{t-1}(H_{t-1}(v_i)) / \tau)
\;\middle\|\;
\operatorname{softmax}(C_{t}(H_{t}(v_i)) / \tau)
\right),
\end{equation}

where $\tau$ is a temperature parameter, and $\tau^2$ is used to compensate for the $1/\tau^2$ scaling of the gradients.
% Unlike hard labels, these soft predictions contain relative information among classes, which helps transfer the knowledge learned from previous tasks.
For the current task, the final objective for graph knowledge preservation is:
\begin{equation}
    \mathcal{L}_{KP} = \alpha_E\mathcal{L}_{E} + \alpha_L\mathcal{L}_{L},
\end{equation}
where $\alpha_E$ and $\alpha_L$ are loss weights.
% The first term constrains the feature space, while the second term preserves the prediction of the previous model.
The whole training process is summarized in Algorithm~\ref{alg:flow_replay}.

\section{Experiments}
\label{sec:experiments}
\subsection{Experimental Setup}
\label{sec:setup}
\textbf{Datasets and Implementation Details.}
We evaluate our method on four real-world datasets: CoraFull~\cite{bojchevski2018deep}, CS~\cite{shchur2018pitfalls}, WikiCS~\cite{mernyei2020wiki}, and Photo~\cite{shchur2018pitfalls}.
More detailed statistics are provided in Appendix~\ref{app:data}.
Following \cite{zhang2022cglb}, we divide the classes into a sequence of tasks for each dataset, and split the nodes in each task into $60$\%, $20$\%, and $20$\% for training, validation, and testing, respectively. 
We add symmetric noise to a fraction of labels in the training set while keeping the validation and test sets clean. 
For each selected node, its label is replaced by a label uniformly sampled from the other classes within the same task.
We also report the results under pair flipping label noise in Appendix~\ref{app:pair}, where each selected label is flipped to a predefined paired class within the same task.
The implementation and detailed settings are provided in Appendix~\ref{app:setting}.

\textbf{Baselines.}
Since there is no existing method specifically designed for our problem, we summarize eight baselines into two categories. 
The first category includes well-known continual learning methods, such as LwF~\cite{li2017learning}, ERGNN~\cite{zhou2021overcoming}, DSLR~\cite{choi2024dslr}, and DMSG~\cite{qiao2025towards}. 
To adapt these methods to noisy labels, we introduce the SCE~\cite{wang2019symmetric} loss during training. 
The second category consists of methods designed for noisy label learning, including SCE, CLNode~\cite{wei2023clnode}, TSS~\cite{wu2024mitigating}, and TFR~\cite{wang2025learning}. 
Since these methods are not originally designed for continual learning, we adopt the CM strategy from ERGNN to construct a memory buffer to prevent catastrophic forgetting. 
For consistency, we store 10 samples per class and use a two-layer GCN as the backbone.
Moreover, two standard reference settings are included.
Joint is an offline upper-bound reference with access to data from all tasks,
while bare model is the backbone GCN without additional strategies, serving as a lower-bound reference.
A full description of the baseline implementations is included in Appendix~\ref{app:Baselines}.

\textbf{Evaluation Metrics.}
To evaluate performance under varying noise levels in continual learning, following previous works \cite{zhang2022cglb,wu2024enhancing,zhu2025federated}, we use average accuracy and average forgetting as the primary metrics. 
Accuracy measures the average accuracy over tasks after learning the final task, 
and Forgetting measures the final retention of previous tasks.
A higher accuracy or a higher forgetting suggests that the model has a better performance.
The formal definitions of metrics are given in Appendix~\ref{app:Metrics}.

\begin{table}[t]
  \centering
  \footnotesize
  \caption{Average performance comparison on  each dataset under noise ratios of $\{0\%, 15\%, 30\%\}$. The best performance is in \textcolor{red}{\textbf{red}}, and the second best is in \textcolor{blue}{\underline{blue}}.}
  \label{tab:comparison}
  \begin{tabular}{lrrrrrr}
    \toprule     
    \multirow{2}{*}{\textbf{Models}} & \multicolumn{3}{c}{\textbf{Accuracy$\uparrow$}} & \multicolumn{3}{c}{\textbf{Forgetting$\uparrow$}} \\
    & $\mathbf{0\%}$ & $\mathbf{15\%}$ & $\mathbf{30\%}$ & $\mathbf{0\%}$ & $\mathbf{15\%}$ & $\mathbf{30\%}$ \\
    \midrule
    \rowcolor{green!20} \multicolumn{7}{c}{\textbf{CoraFull}} \\
    Joint & $90.40_{\pm0.10}$ & $75.82_{\pm0.12}$ & $62.45_{\pm0.27}$ & \multicolumn{3}{c}{-} \\
    Bare model & $34.21_{\pm1.69}$ & $31.32_{\pm6.43}$ & $30.60_{\pm1.13}$ & $-61.91_{\pm3.50}$ & $-49.15_{\pm9.88}$ & $-36.00_{\pm1.09}$ \\
    LwF & $60.32_{\pm0.77}$ & $55.25_{\pm2.14}$ & $50.87_{\pm1.54}$ & $-33.41_{\pm0.95}$ & $-24.84_{\pm2.82}$ & $-15.57_{\pm1.43}$ \\ 
    ERGNN & $77.74_{\pm0.15}$ & $40.91_{\pm0.34}$ & $33.94_{\pm0.58}$ & \textcolor{red}{$\mathbf{-4.62_{\pm0.37}}$} & $-33.06_{\pm0.40}$ & $-32.94_{\pm0.37}$ \\
    DSLR & $74.23_{\pm0.90}$ & $62.09_{\pm1.62}$ & $51.06_{\pm0.99}$ & $-14.47_{\pm0.84}$ & $-11.75_{\pm1.46}$ & $-8.98_{\pm0.75}$ \\
    DMSG & \textcolor{blue}{\underline{$78.04_{\pm0.56}$}} & \textcolor{blue}{\underline{$67.20_{\pm0.82}$}} & \textcolor{blue}{\underline{$59.71_{\pm1.30}$}} & $-13.20_{\pm0.66}$ & \textcolor{blue}{\underline{$-10.65_{\pm0.81}$}} & \textcolor{blue}{\underline{$-7.68_{\pm0.85}$}} \\
    SCE & $39.30_{\pm0.15}$ & $41.27_{\pm1.21}$ & $34.41_{\pm1.77}$ & $-56.65_{\pm0.28}$ & $-42.06_{\pm1.45}$ & $-35.08_{\pm2.14}$ \\
    CLNode & $78.01_{\pm0.18}$ & $30.12_{\pm0.81}$ & $27.13_{\pm0.22}$ & \textcolor{blue}{\underline{$-5.06_{\pm0.31}$}} & $-53.95_{\pm0.37}$ & $-55.91_{\pm0.52}$ \\
    TSS & $75.02_{\pm0.57}$ & $27.61_{\pm1.89}$ & $25.27_{\pm1.69}$ & $-9.20_{\pm1.30}$ & $-52.97_{\pm1.22}$ & $-47.19_{\pm1.38}$ \\
    TFR & $49.08_{\pm0.61}$ & $36.08_{\pm1.35}$ & $26.89_{\pm1.36}$ & $-45.95_{\pm0.72}$ & $-45.92_{\pm0.61}$ & $-42.35_{\pm1.09}$ \\
    UFO & \textcolor{red}{$\mathbf{81.78_{\pm0.85}}$} & \textcolor{red}{$\mathbf{80.71_{\pm0.08}}$} & \textcolor{red}{$\mathbf{77.77_{\pm1.04}}$} & $-6.35_{\pm1.10}$ & \textcolor{red}{$\mathbf{-6.33_{\pm2.25}}$} & \textcolor{red}{$\mathbf{-7.10_{\pm1.27}}$} \\
    
    \rowcolor{blue!20} \multicolumn{7}{c}{\textbf{CS}} \\
    Joint & $97.00_{\pm0.10}$ & $67.43_{\pm4.15}$ & $51.06_{\pm1.83}$ & \multicolumn{3}{c}{-} \\
    Bare model & $61.22_{\pm0.10}$ & $62.83_{\pm0.07}$ & $53.04_{\pm0.03}$ & $-45.55_{\pm0.02}$ & $-28.98_{\pm0.01}$ & $-25.10_{\pm0.21}$ \\
    LwF & $92.51_{\pm4.10}$ & $53.39_{\pm6.40}$ & $45.19_{\pm15.55}$ & $-5.49_{\pm4.78}$ & $-40.94_{\pm8.02}$ & $-32.99_{\pm13.22}$ \\
    ERGNN & $94.40_{\pm0.14}$ & $57.82_{\pm2.86}$ & $47.73_{\pm3.43}$ & $-3.26_{\pm0.29}$ & $-20.09_{\pm6.03}$ & $-11.67_{\pm0.85}$ \\
    DSLR & \textcolor{blue}{\underline{$96.47_{\pm0.80}$}} & \textcolor{blue}{\underline{$77.67_{\pm0.45}$}} & \textcolor{blue}{\underline{$68.10_{\pm1.80}$}} & \textcolor{blue}{\underline{$-2.01_{\pm0.88}$}} & $-12.98_{\pm1.59}$ & $-2.29_{\pm2.39}$ \\
    DMSG & $93.87_{\pm1.14}$ & $70.46_{\pm3.67}$ & $57.55_{\pm8.96}$ & $-4.24_{\pm1.43}$ & \textcolor{blue}{\underline{$-6.34_{\pm4.45}$}} & \textcolor{red}{$\mathbf{7.13_{\pm9.74}}$} \\
    SCE & $78.90_{\pm7.20}$ & $58.42_{\pm8.09}$ & $46.00_{\pm6.30}$ & $-22.48_{\pm8.69}$ & $-33.80_{\pm11.78}$ & $-29.68_{\pm5.30}$ \\
    CLNode & $93.84_{\pm0.32}$ & $51.80_{\pm1.52}$ & $57.89_{\pm1.22}$ & $-3.84_{\pm0.47}$ & $-55.76_{\pm1.90}$ & $-47.60_{\pm1.36}$ \\
    TSS & $79.62_{\pm5.16}$ & $55.28_{\pm9.70}$ & $48.12_{\pm6.16}$ & $-14.29_{\pm8.06}$ & $-41.88_{\pm12.52}$ & $-36.27_{\pm8.29}$ \\
    TFR & $94.11_{\pm1.98}$ & $28.33_{\pm9.63}$ & $28.81_{\pm8.76}$ & $-3.43_{\pm2.58}$ & $-48.21_{\pm15.95}$ & $-11.52_{\pm6.81}$ \\
    UFO & \textcolor{red}{$\mathbf{97.29_{\pm0.29}}$} & \textcolor{red}{$\mathbf{92.21_{\pm0.43}}$} & \textcolor{red}{$\mathbf{81.46_{\pm0.43}}$} & \textcolor{red}{$\mathbf{-1.30_{\pm0.27}}$} & \textcolor{red}{$\mathbf{0.06_{\pm1.34}}$} & \textcolor{blue}{\underline{$3.14_{\pm0.20}$}} \\

    \rowcolor{yellow!20} \multicolumn{7}{c}{\textbf{WikiCS}} \\
    Joint & $92.70_{\pm0.40}$ & $90.88_{\pm0.40}$ & $86.41_{\pm3.01}$ & \multicolumn{3}{c}{-} \\
    Bare model & $77.72_{\pm4.51}$ & $68.95_{\pm6.56}$ & $65.24_{\pm7.98}$ & $-17.67_{\pm5.91}$ & $-21.10_{\pm3.59}$ & $-15.06_{\pm11.64}$ \\
    LwF & \textcolor{blue}{\underline{$88.75_{\pm1.54}$}} & $74.06_{\pm4.37}$ & $62.76_{\pm4.96}$ & \textcolor{blue}{\underline{$-4.72_{\pm2.09}$}} & $-20.50_{\pm6.35}$ & $-20.08_{\pm6.83}$ \\
    ERGNN & $86.58_{\pm0.36}$ & $52.77_{\pm3.69}$ & $48.62_{\pm6.77}$ & $-6.54_{\pm0.37}$ & $-44.23_{\pm10.28}$ & $-39.28_{\pm2.36}$ \\
    DSLR & $82.71_{\pm0.71}$ & $70.35_{\pm2.87}$ & $61.74_{\pm1.39}$ & $-8.65_{\pm0.44}$ & $-12.93_{\pm2.34}$ & \textcolor{blue}{\underline{$-8.49_{\pm2.93}$}} \\
    DMSG & $83.69_{\pm1.55}$ & \textcolor{blue}{\underline{$82.18_{\pm2.34}$}} & \textcolor{blue}{\underline{$66.11_{\pm5.12}$}} & $-10.84_{\pm1.48}$ & \textcolor{blue}{\underline{$-9.69_{\pm2.01}$}} & $-18.36_{\pm10.72}$ \\
    SCE & $76.26_{\pm0.40}$ & $65.59_{\pm2.69}$ & $52.77_{\pm3.74}$ & $-21.16_{\pm0.81}$ & $-29.05_{\pm4.66}$ & $-24.16_{\pm4.64}$ \\
    CLNode & $80.25_{\pm2.05}$ & $52.31_{\pm1.36}$ & $55.63_{\pm2.62}$ & $-14.70_{\pm2.43}$ & $-49.74_{\pm1.13}$ & $-45.26_{\pm2.78}$ \\
    TSS & $73.99_{\pm0.55}$ & $52.14_{\pm6.47}$ & $49.80_{\pm6.78}$ & $-13.69_{\pm1.05}$ & $-42.47_{\pm7.24}$ & $-40.11_{\pm10.63}$ \\
    TFR & $83.37_{\pm0.82}$ & $60.88_{\pm7.36}$ & $42.42_{\pm2.64}$ & $-8.04_{\pm3.94}$ & $-14.95_{\pm10.51}$ & $-35.46_{\pm1.84}$ \\
    UFO & \textcolor{red}{$\mathbf{93.36_{\pm0.01}}$} & \textcolor{red}{$\mathbf{93.29_{\pm0.08}}$} & \textcolor{red}{$\mathbf{83.10_{\pm0.89}}$} & \textcolor{red}{$\mathbf{-0.69_{\pm0.07}}$} & \textcolor{red}{$\mathbf{-0.66_{\pm0.10}}$} & \textcolor{red}{$\mathbf{-8.27_{\pm1.16}}$} \\

    \rowcolor{red!20} \multicolumn{7}{c}{\textbf{Photo}} \\
    Joint & $96.90_{\pm0.80}$ & $93.24_{\pm0.72}$ & $85.87_{\pm2.34}$ & \multicolumn{3}{c}{-} \\
    Bare model & $65.75_{\pm0.01}$ & $49.64_{\pm0.02}$ & $49.63_{\pm0.05}$ & $-38.78_{\pm0.07}$ & $-30.02_{\pm0.04}$ & $-22.34_{\pm0.06}$ \\
    LwF & $80.03_{\pm19.44}$ & $79.21_{\pm9.37}$ & $63.36_{\pm1.30}$ & $-18.51_{\pm18.04}$ & $-12.05_{\pm19.43}$ & $-17.23_{\pm16.98}$ \\
    ERGNN & $95.49_{\pm0.14}$ & $42.62_{\pm11.31}$ & $53.00_{\pm17.22}$ & $-5.37_{\pm7.26}$ & \textcolor{blue}{\underline{$-0.48_{\pm1.65}$}} & $-30.88_{\pm11.94}$ \\
    DSLR & \textcolor{blue}{\underline{$97.36_{\pm0.14}$}} & \textcolor{blue}{\underline{$87.70_{\pm1.60}$}} & \textcolor{blue}{\underline{ $69.62_{\pm3.95}$}} & $-1.28_{\pm0.36}$ & $-11.90_{\pm2.64}$ & $-18.56_{\pm8.21}$ \\
    DMSG & $49.58_{\pm0.02}$ & $57.11_{\pm7.66}$ & $50.27_{\pm0.89}$ & \textcolor{blue}{\underline{$-0.31_{\pm0.15}$}} & $-23.91_{\pm12.93}$ & \textcolor{blue}{\underline{$0.77_{\pm1.24}$}} \\
    SCE & $63.85_{\pm5.86}$ & $47.63_{\pm5.82}$ & $57.59_{\pm11.41}$ & $-42.02_{\pm7.47}$ & $-37.85_{\pm20.60}$ & $-10.61_{\pm15.30}$ \\
    CLNode & $89.31_{\pm9.31}$ & $57.28_{\pm11.90}$ & $59.86_{\pm4.83}$ & $-8.53_{\pm11.61}$ & $-50.99_{\pm14.67}$ & $-46.61_{\pm6.51}$ \\
    TSS & $61.44_{\pm11.38}$ & $62.42_{\pm8.31}$ & $55.51_{\pm10.22}$ & $-34.36_{\pm19.57}$ & $-27.04_{\pm20.64}$ & $-35.44_{\pm4.19}$ \\
    TFR & $73.87_{\pm4.70}$ & $51.82_{\pm4.04}$ & $46.42_{\pm1.78}$ & $-28.18_{\pm6.61}$ & $-28.01_{\pm12.53}$ & $-32.90_{\pm1.15}$ \\
    UFO & \textcolor{red}{$\mathbf{98.07_{\pm0.04}}$} & \textcolor{red}{$\mathbf{96.04_{\pm0.25}}$} & \textcolor{red}{$\mathbf{93.40_{\pm0.86}}$} & \textcolor{red}{$\mathbf{-0.22_{\pm0.06}}$} & \textcolor{red}{$\mathbf{-0.17_{\pm0.49}}$} & \textcolor{red}{$\mathbf{1.61_{\pm0.99}}$} \\
    \bottomrule
  \end{tabular}
\end{table}
% \vspace{-1.8em}
\subsection{Comparison Study}
Table~\ref{tab:comparison} shows that UFO performs better on four datasets under varying  noise ratios.
As the noise ratio increases, all models show lower average accuracy, while forgetting first decreases and then increases.
This is because noise may already hurt the model at an early stage, preventing it from learning true knowledge well, so the later performance change becomes less obvious.
Existing CGL methods mainly focus on mitigating catastrophic forgetting, so their average forgetting remains relatively stable.
On CoraFull, when the noise ratio increases from 0\% to 30\%, the accuracy of DSLR decreases by 23.17\%, while its forgetting increases by 5.49\%.
However, among graph learning with label noise methods, the forgetting of CLNode and TSS decreases by 50.85\% and 37.99\%.
This may be related to its higher early accuracy, and continuously arriving noisy labels further affect the stability of the model.
In contrast, UFO shows smaller changes in both accuracy and forgetting on all four datasets, demonstrating stronger robustness under noisy conditions.

\begin{figure}[h]
\centering
% 图 1
\begin{subfigure}{0.18\linewidth}
    \centering
    \includegraphics[width=\linewidth]{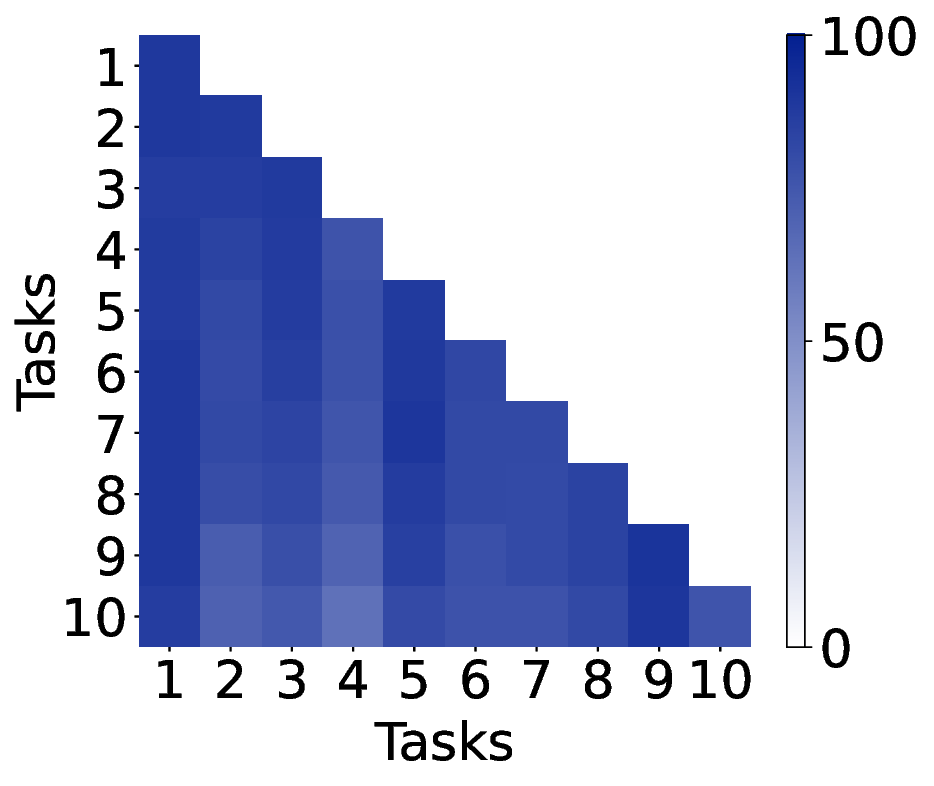}
    \caption{UFO}
\end{subfigure}
\hfill
\begin{subfigure}{0.18\linewidth}
    \centering
    \includegraphics[width=\linewidth]{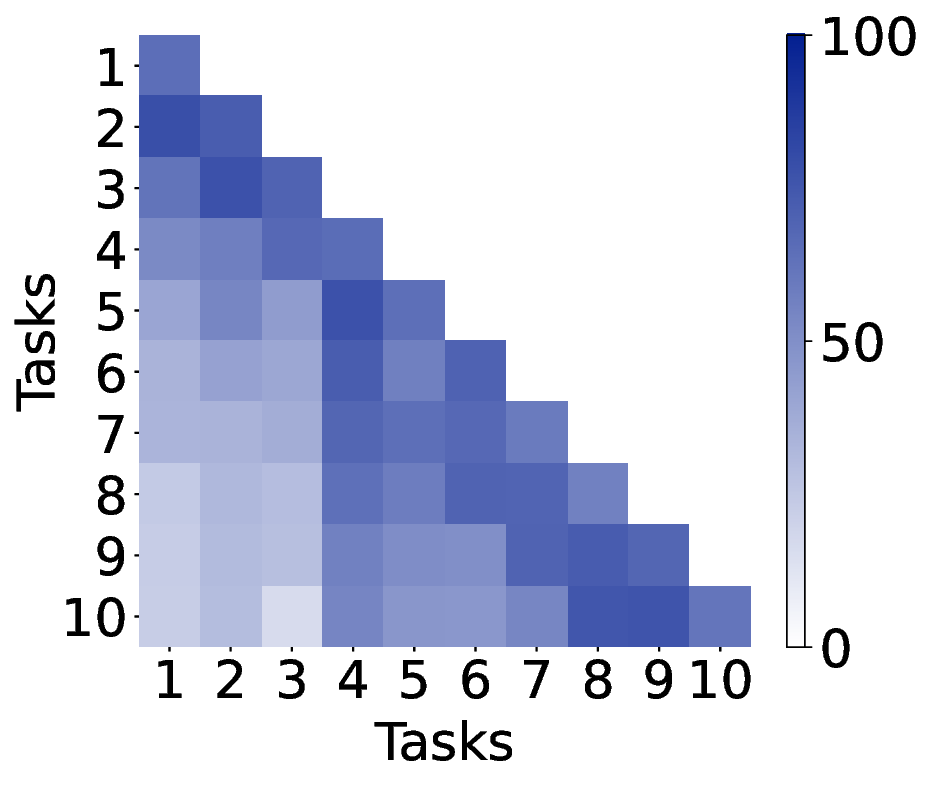}
    \caption{LwF}
\end{subfigure}
\hfill
\begin{subfigure}{0.18\linewidth}
    \centering
    \includegraphics[width=\linewidth]{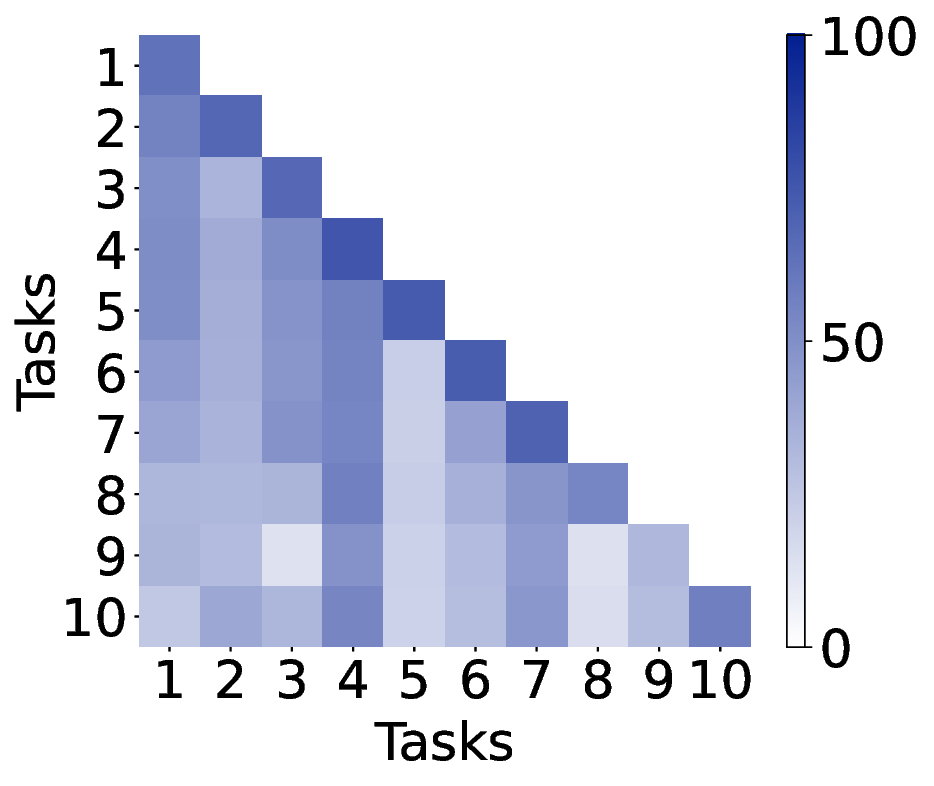}
    \caption{ERGNN}
\end{subfigure}
\hfill
\begin{subfigure}{0.18\linewidth}
    \centering
    \includegraphics[width=\linewidth]{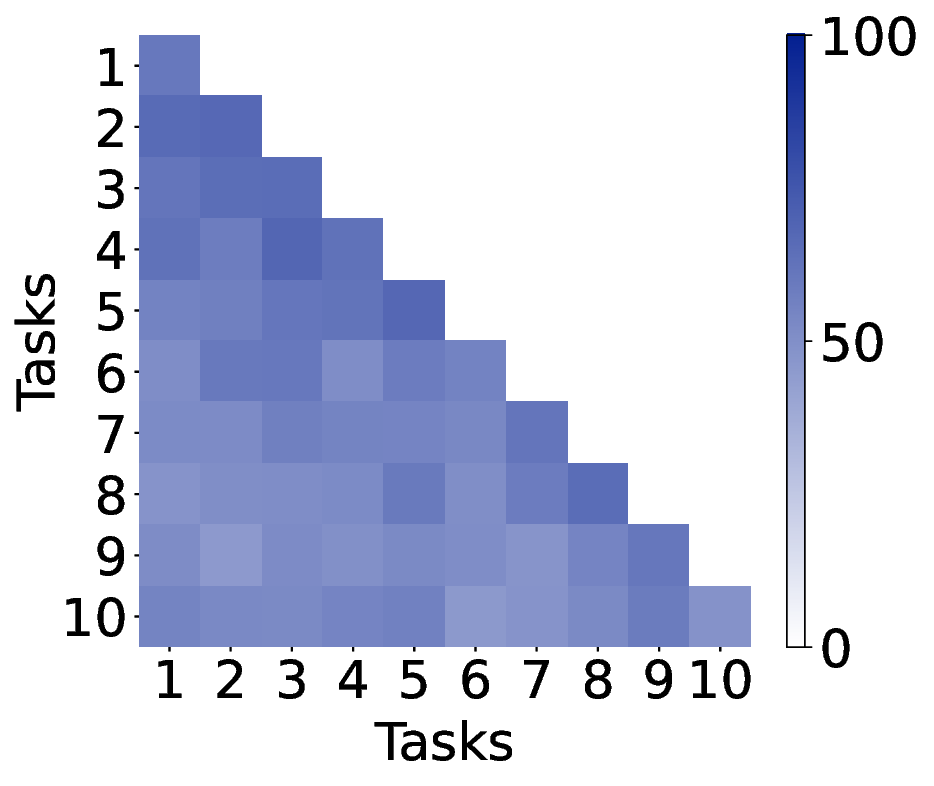}
    \caption{DSLR}
\end{subfigure}
\hfill
\begin{subfigure}{0.18\linewidth}
    \centering
    \includegraphics[width=\linewidth]{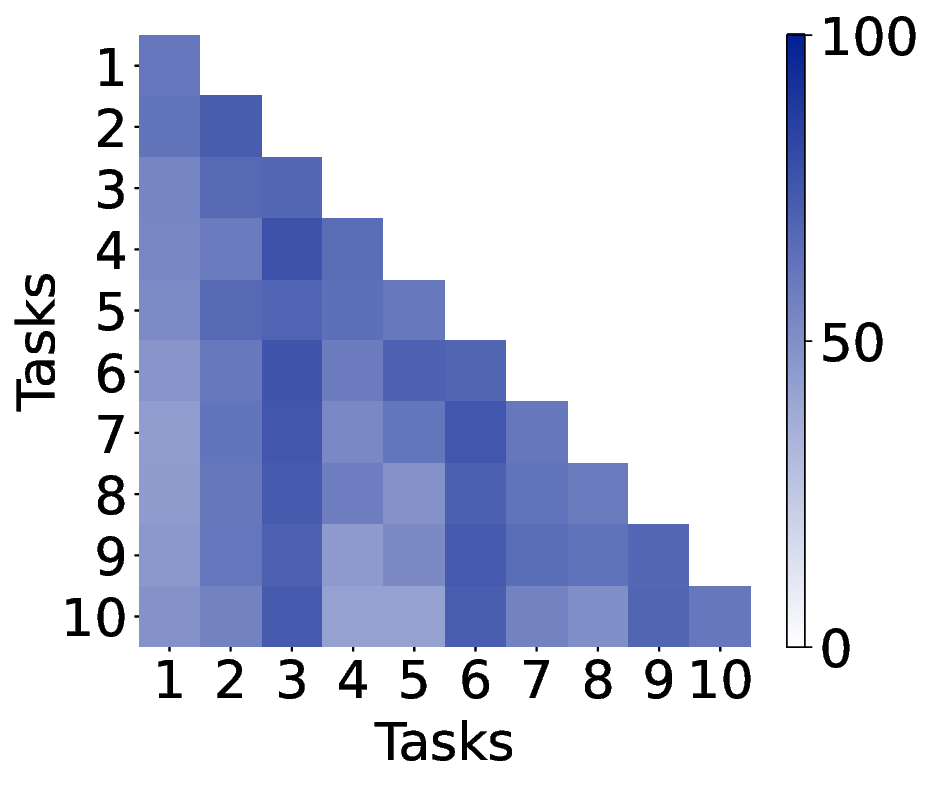}
    \caption{DMSG}
\end{subfigure}
\caption{Performance matrices on CoraFull with $30$\% noise.}
\label{fig:continualmethod}
\end{figure}

To further analyze the overall behavior of different models, we visualize the performance matrices on CoraFull under the 30\% noise setting.
As shown in Figure~\ref{fig:continualmethod}, darker colors indicate higher accuracy.
For LwF and ERGNN, the dark regions are concentrated along the diagonal and the colors become lighter for earlier tasks.
% This indicates that they perform well mainly on recently learned tasks and struggle with  the noisy labels.
Under noisy labels, DSLR and DMSG show relatively uniform but lighter color distributions.
In contrast, UFO shows a more stable and darker matrix, demonstrating that it not only preserves knowledge but also reduces
the accumulation of corrupted supervision,
which achieves a better balance between catastrophic forgetting and catastrophic remembering.
Due to space limitations, more comparisons are provided in Appendix~\ref{app:visualization}.

\vspace{-0.5em}
\subsection{Ablation Study}
To further analyze the effectiveness of each component, we conduct an ablation study on CoraFull and CS with different noise ratios. 
We consider the bare model (BM), knowledge preservation (KP), relative scores for the new task (NS) and old task (OS), and the replay mechanism (R).
The results in Table~\ref{tab:ablation_corafull_transposed} show that:
(1) KP enables the BM to achieve better performance by preserving previous knowledge and reducing feature drift. 
(2) When NS is further introduced, the model achieves higher accuracy in several noise ratios, especially under 15\% noise. 
(3) R brings clear gains, e.g., improving the forgetting value on CoraFull with $30\%$ noise from $-53.92\%$ to $-7.93\%$.
(4) After adding OS, UFO achieves the best overall performance across varying noise levels.
This suggests that OS helps filter low-quality features generated by the flow model.

\begin{table}[h]
    \centering
    \footnotesize
    \renewcommand{\arraystretch}{1.2}
    \setlength{\tabcolsep}{4pt}
    \caption{Ablation study of our UFO. The best performance is highlighted in \textcolor{red}{\textbf{red}}.}
    \label{tab:ablation_corafull_transposed}
    \begin{tabular}{lrrrrrr}
        \hline
        \textbf{Metric} & \textbf{Noise} & \textbf{BM} & \textbf{BM+KP} & \textbf{BM+KP+NS} & \textbf{BM+KP+NS+R} & \textbf{UFO} \\
        \hline
        \rowcolor{green!20} \multicolumn{7}{c}{\textbf{CoraFull}} \\
        \multirow{3}{*}{\textbf{Accuracy$\uparrow$}} 
        & $0\%$ & $33.71_{\pm4.74}$ & $47.14_{\pm0.40}$ & $47.07_{\pm0.60}$ & $53.16_{\pm2.48}$ & \textcolor{red}{$\mathbf{81.78_{\pm0.85}}$} \\
        & $15\%$ & $34.02_{\pm0.58}$ & $39.19_{\pm0.47}$ & $40.78_{\pm0.49}$ & $51.21_{\pm0.12}$ & \textcolor{red}{$\mathbf{80.71_{\pm0.08}}$} \\
        & $30\%$ & $29.02_{\pm2.34}$ & $33.22_{\pm1.00}$ & $33.27_{\pm0.46}$ & $46.65_{\pm1.01}$ & \textcolor{red}{$\mathbf{77.77_{\pm1.04}}$} \\
        \hline
        \multirow{3}{*}{\textbf{Forgetting$\uparrow$}} 
        & $0\%$ & $-62.09_{\pm5.34}$ & $-47.37_{\pm0.46}$ & $-47.39_{\pm0.71}$ & $-6.39_{\pm2.32}$ & \textcolor{red}{$\mathbf{-6.35_{\pm1.10}}$} \\
        & $15\%$ & $-47.04_{\pm0.37}$ & $-53.46_{\pm0.62}$ & $-51.78_{\pm0.61}$ & $-6.57_{\pm0.79}$ & \textcolor{red}{$\mathbf{-6.33_{\pm2.25}}$} \\
        & $30\%$ & $-38.10_{\pm2.53}$ & $-53.98_{\pm1.17}$ & $-53.92_{\pm0.55}$ & $-7.93_{\pm1.08}$ & \textcolor{red}{$\mathbf{-7.10_{\pm1.27}}$} \\
        \hline
        \rowcolor{blue!20} \multicolumn{7}{c}{\textbf{CS}} \\
        \multirow{3}{*}{\textbf{Accuracy$\uparrow$}} 
        & $0\%$ & $61.22_{\pm0.10}$ & $93.53_{\pm0.54}$ & $92.31_{\pm0.48}$ & $96.16_{\pm0.51}$ & \textcolor{red}{$\mathbf{97.29_{\pm0.29}}$} \\
        & $15\%$ & $62.83_{\pm0.07}$ & $87.15_{\pm1.10}$ & $89.18_{\pm0.36}$ & $90.32_{\pm1.12}$ & \textcolor{red}{$\mathbf{92.21_{\pm0.43}}$} \\
        & $30\%$ & $53.04_{\pm0.03}$ & $69.91_{\pm3.80}$ & $67.11_{\pm3.60}$ & $78.65_{\pm3.44}$ & \textcolor{red}{$\mathbf{81.46_{\pm0.43}}$} \\
        \hline
        \multirow{3}{*}{\textbf{Forgetting$\uparrow$}} 
        & $0\%$ & $-45.55_{\pm0.02}$ & $-5.59_{\pm0.62}$ & $-7.09_{\pm0.59}$ & $-2.36_{\pm0.60}$ & \textcolor{red}{$\mathbf{-1.30_{\pm0.27}}$} \\
        & $15\%$ & $-28.98_{\pm0.01}$ & $-1.99_{\pm1.25}$ & $0.29_{\pm0.58}$ & $-0.22_{\pm1.22}$ & \textcolor{red}{$\mathbf{0.06_{\pm1.34}}$} \\
        & $30\%$ & $-25.10_{\pm0.21}$ & $-8.14_{\pm4.58}$ & $-12.28_{\pm4.20}$ & $-0.32_{\pm4.19}$ & \textcolor{red}{$\mathbf{3.14_{\pm0.20}}$} \\
        \hline
    \end{tabular}
\end{table}

\subsection{Hyperparameter}
\begin{figure}[h]
    \centering
        \includegraphics[width=\textwidth]{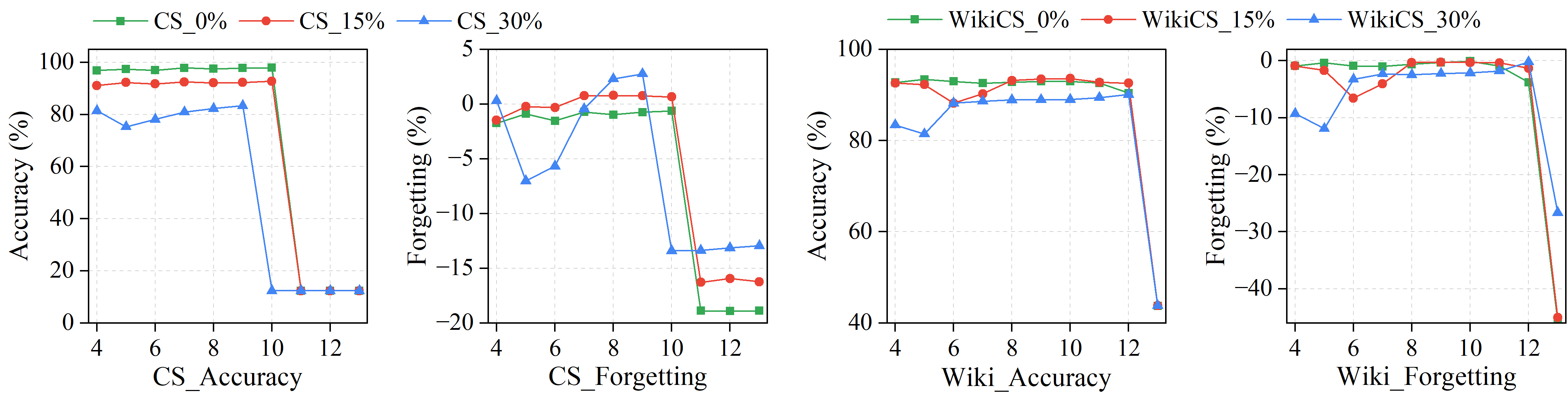}
    \caption{Sensitivity analysis for the number of flows with 0\%, 15\%, and 30\% noise.}
\label{fig:Hyperparameter}
\end{figure}

We evaluate the sensitivity to the number of flows in Figure~\ref{fig:Hyperparameter}.
As observed, the performance remains stable with $0\%$ noise, indicating a shallow or moderate flow is sufficient for clean distributions.
When the noise ratio increases to $30\%$, accuracy and forgetting reach their best levels around 9--12 flows.
With the number of flows further increasing to 13, the results drop clearly.
This suggests that a moderate flow depth provides a better balance for robust continual learning, while an overly deep flow increases optimization difficulty.

\subsection{Visualization}
\begin{figure}[htbp]
    \centering
    \includegraphics[width=\linewidth]{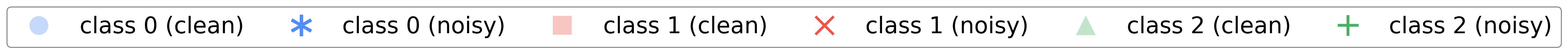}\\
    \vspace{3pt}

    \begin{subfigure}{0.32\linewidth}
        \centering
        \includegraphics[width=\linewidth]{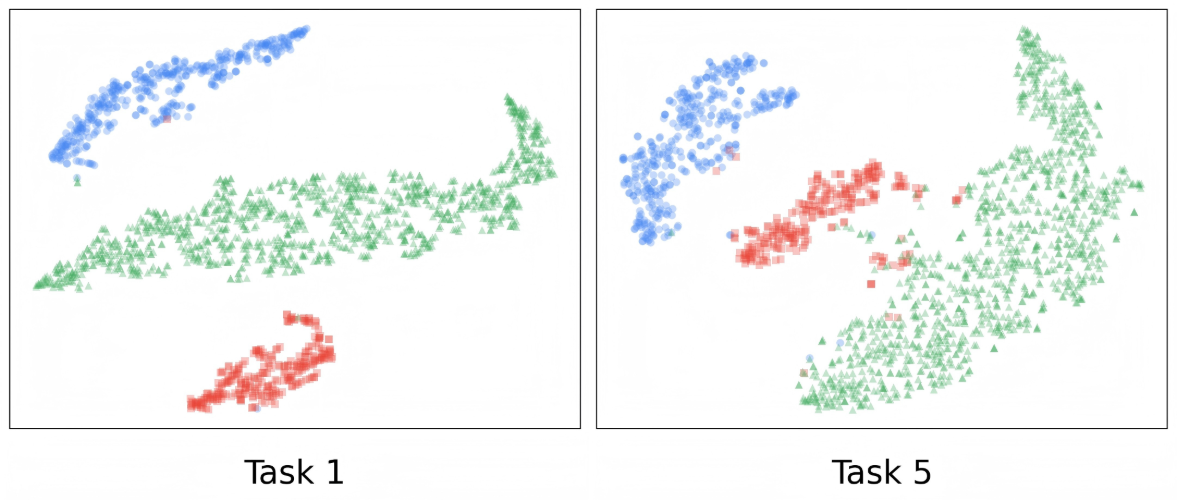}
        \caption{UFO with 0\% noise}
    \end{subfigure}
    \hfill
    \begin{subfigure}{0.32\linewidth}
        \centering
        \includegraphics[width=\linewidth]{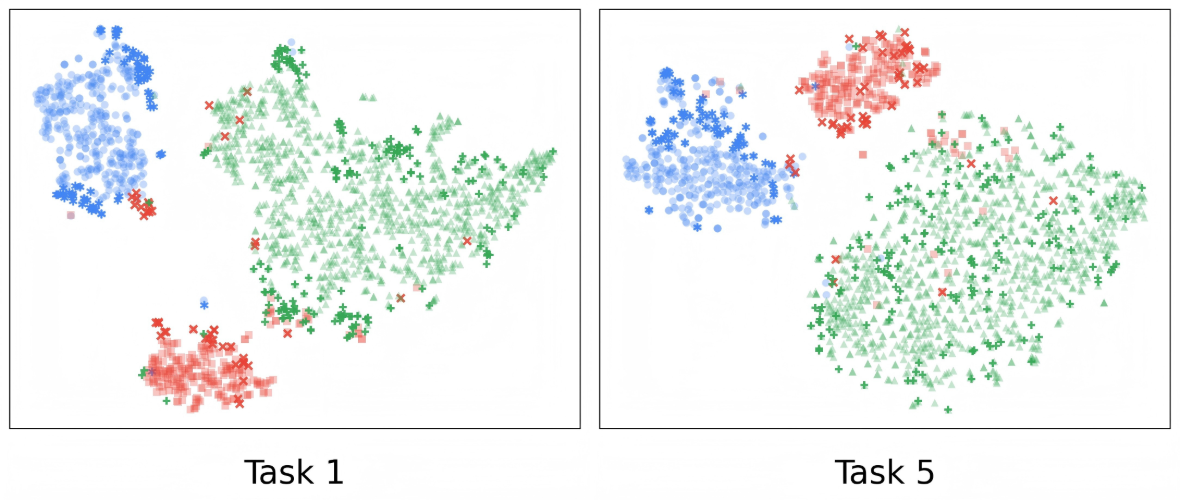}
        \caption{UFO with 15\% noise}
    \end{subfigure}
    \hfill
    \begin{subfigure}{0.32\linewidth}
        \centering
        \includegraphics[width=\linewidth]{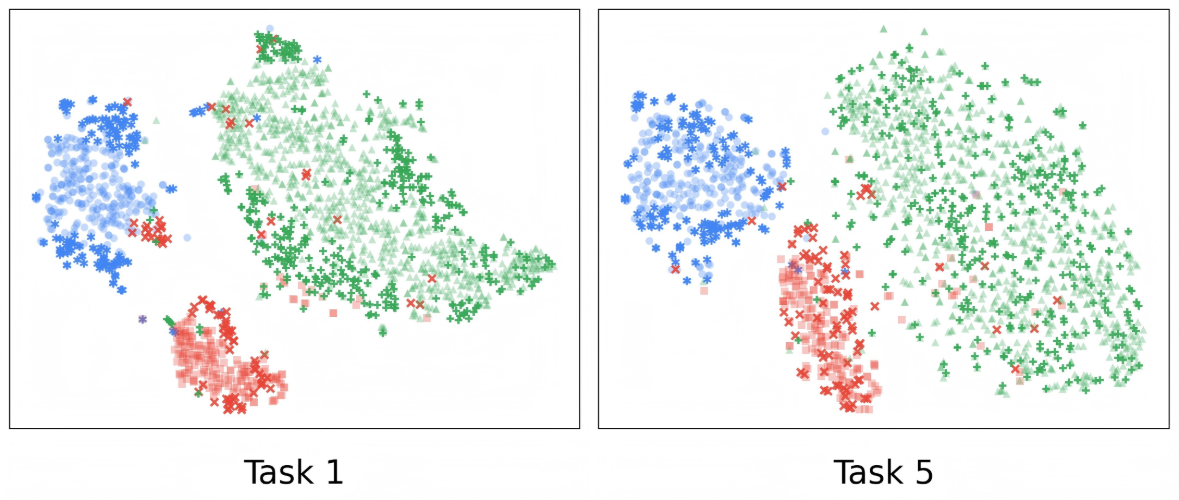}
        \caption{UFO with 30\% noise}
    \end{subfigure}

    \caption{t-SNE visualization of the evolution of embeddings for classes 0, 1, and 2 on sequential CS.}
    \label{fig:t_SNE111}
\end{figure}

To analyze the evolution of clean and noisy nodes, 
we visualize the embeddings of the classes from the first task (classes 0, 1, and 2) on CS in Figure~\ref{fig:t_SNE111}.
The three subfigures show the embeddings learned by UFO on sequential CS with 0\%, 15\%, and 30\% noise.
The 0\% noise setting provides a clean reference, where the embeddings form distinct clusters with minimal overlap. 
As the noise ratio increases to 15\%,30\%, we can observe that noisy nodes in UFO are mainly located near the boundary or the outer region of each cluster.
Comparing Task 1 and Task 5,
the embeddings of old classes show less drift and preserve relatively stable structures.
This suggests that UFO can preserve knowledge and maintain robust representations under noisy supervision.
The complete visualization results for UFO and Joint are provided in Appendix~\ref{ref:t-sne-joint}.

% \vspace{-1em}
\section{Limitations}
\label{sec:limitations}
This work has two limitations.
First, UFO performs replay in the feature space and does not explicitly reconstruct the original graph topology, such as historical edges and neighborhoods.
This may limit the ability of feature replay to fully recover structural knowledge from previous tasks.
Second, UFO relies on the configuration of the flow model.
An overly deep flow can increase optimization difficulty and computational cost.
Thus, the number of flow layers and related hyperparameters need to be selected carefully in practice.
\section{Conclusion}
\label{sec:conclusion}
In this paper, we study a more realistic robust continual graph learning scenario and identify a new failure mode caused by the accumulation of noisy supervision.
To jointly address these challenges, we propose UFO, a unified flow-oriented framework.
UFO uses normalizing flows to conditionally model feature distributions for replay and estimate instance-level reliability under corrupted supervision.
Furthermore, we introduce graph knowledge preservation to constrain structural and semantic consistency.
Experimental results demonstrate the superiority of UFO in improving performance and robustness under varying noise levels.

\bibliographystyle{unsrt}
\bibliography{ref}
\newpage
%%%%%%%%%%%%%%%%%%%%%%%%%%%%%%%%%%%%%%%%%%%%%%%%%%%%%%%%%%%%
\appendix
% \section{Overview}
\section{Implementation Details}
\label{app:implementation}
\subsection{Datasets} 
\label{app:data}
In our experiments, we evaluate our method on four widely used graph datasets.
\begin{itemize} [leftmargin=0.5cm]
    \item \textbf{CoraFull} \cite{bojchevski2018deep} is an academic citation network with 70 classes, where nodes denote academic papers and edges denote citation links between papers. 
    \item \textbf{CS} \cite{shchur2018pitfalls} is an author collaboration network extracted from the Microsoft Academic Graph. In this dataset, each node represents an author, and an edge connects two authors who have co-authored a paper.
    \item \textbf{WikiCS} \cite{mernyei2020wiki} is built from Wikipedia on computer science topics. It treats articles as nodes and hyperlinks as edges, and provides multiple predefined splits for reliable evaluation.
    \item \textbf{Photo} \cite{shchur2018pitfalls} is an Amazon co-purchase network from the photography category. Nodes represent products, and an edge is created when two items are frequently purchased together, reflecting implicit product relationships.
\end{itemize}
The statistics of these datasets are summarized in Table~\ref{tab:dataset_statistics}.
\begin{table}[H]
    \centering
    \footnotesize
    \renewcommand{\arraystretch}{1.2}
    \setlength{\tabcolsep}{12pt}
    \caption{Statistics of the Datasets.}
    \label{tab:dataset_statistics}
    \begin{tabular}{lrrrrr}
        \hline
        \textbf{Dataset} & \textbf{\#Nodes} & \textbf{\#Edges} & \textbf{\#Features} & \textbf{\#Classes} & \textbf{\#Tasks} \\
        \hline
        \textbf{CoraFull} & $19,793$ & $126,842$ & $8,710$ & $70$ & $10$ \\
        \textbf{CS}       & $18,333$ & $163,788$ & $6,805$ & $15$ & $5$  \\
        \textbf{WikiCS}   & $11,701$ & $431,726$ & $300$   & $10$ & $5$  \\
        \textbf{Photo}    & $7,650$  & $238,162$ & $745$   & $8$  & $4$  \\
        \hline
    \end{tabular}
\end{table}

\subsection{Baselines}
\label{app:Baselines}
We compare our method with several state-of-the-art approaches from both continual graph learning and graph learning with label noise. 
The comparison methods are briefly introduced as follows.
\begin{itemize} [leftmargin=0.5cm]
    \item \textbf{Joint} trains the backbone GCN on all tasks simultaneously.
    Since it has access to the full task sequence, it is reported only as an upper-bound reference.
    This setting removes the influence of catastrophic forgetting and highlights the negative effect of noisy supervision.
    \item \textbf{Bare model} is the backbone GCN without any continual learning or robust strategy.
    It serves as a lower-bound reference, where the model is directly affected by catastrophic forgetting and catastrophic remembering.
\end{itemize}

\par Continual graph learning (adding an SCE loss term weighted by $0.5$):
\begin{itemize} [leftmargin=0.5cm]
    \item \textbf{LwF} \cite{li2017learning} is a regularization-based method that preserves knowledge from previous tasks by distilling soft targets from a historical model.
    \item \textbf{ERGNN} \cite{zhou2021overcoming} selects representative nodes from previous tasks and replays them to reduce catastrophic forgetting in graph neural networks.
    \item \textbf{DSLR} \cite{choi2024dslr} improves experience replay by considering coverage-based diversity and graph structure learning, which helps preserve both feature and structural information.
    \item \textbf{DMSG} \cite{qiao2025towards} improves memory replay by combining sample selection with generation, with explicit consideration of memory diversity in continual graph learning.
\end{itemize}
    
\par Graph learning with label noise (replaying $10$ nodes per class):
\begin{itemize} [leftmargin=0.5cm]
    \item \textbf{SCE} \cite{wang2019symmetric} is a robust loss function that combines cross-entropy with reverse cross-entropy, making the model less sensitive to noisy labels.
    \item \textbf{CLNode} \cite{wei2023clnode} uses curriculum learning for graph data, where node difficulty is estimated from multiple perspectives and the model is trained from easier to harder samples.
    \item \textbf{TSS} \cite{wu2024mitigating} leverages graph topology to iteratively select high-confidence nodes, allowing the model to learn from more reliable samples under label noise.
    \item \textbf{TFR} \cite{wang2025learning} introduces a dual-network framework with a backbone GNN and a decoder GNN, reconstructing topological features to enhance robustness via mutual information maximization.
\end{itemize}

\subsection{Evaluation Metrics} \label{app:Metrics}
We adopt two commonly used metrics in continual learning, namely accuracy and forgetting \cite{zhang2022cglb}, to evaluate both the overall performance and robustness of the model across tasks.
Formally, they are defined as:
\begin{equation}
    \mathrm{Accuracy} = \frac{1}{n} \sum_{i=1}^{n} A_{n,i}, \quad
\mathrm{Forgetting} = \frac{1}{n-1} \sum_{i=1}^{n-1} \left( A_{n,i} - A_{i,i} \right),
\end{equation}
where $n$ denotes the total number of tasks, and $A_{i,j}$ represents the test accuracy on task $j$ after training up to task $i$. In particular, $A_{i,i}$ is the accuracy immediately after learning task $i$, while $A_{n,i}$ denotes its final accuracy after all tasks are learned.
\subsection{Details Settings}
\label{app:setting}
We adopt the GCN as the backbone, with a hidden dimension of $256$.
Following \cite{cao2025erroreraser,zhang2022cglb}, the model is trained using the Adam optimizer with a learning rate of 0.005 for 200 epochs on each task.
The flow model consists of 4 affine coupling layers and 4 permutation layers.
The temperature parameter is set to $\tau=2.71$, with $\alpha_E=0.09$ and $\alpha_L=0.06$.
All results are reported as the mean and standard deviation over three independent runs, with implementation based on PyTorch 2.3.0 and CUDA 12.1 on a 48GB vGPU.

\begin{table}[H]
  \centering
  \footnotesize
  \caption{Average performance comparison on CS under pair flipping noise.
  The best performance is in \textcolor{red}{\textbf{red}}, and the second best is in \textcolor{blue}{\underline{blue}}.}
  \label{tab:dataset_subscript_pm}
  \begin{tabular}{lrrrrrr}
    \toprule
    \multirow{2}{*}{\textbf{Models}} & \multicolumn{3}{c}{\textbf{Accuracy$\uparrow$}} & \multicolumn{3}{c}{\textbf{Forgetting$\uparrow$}} \\
    & $\mathbf{0\%}$ & $\mathbf{15\%}$ & $\mathbf{30\%}$ & $\mathbf{0\%}$ & $\mathbf{15\%}$ & $\mathbf{30\%}$ \\
    \midrule
    \rowcolor{blue!20} \multicolumn{7}{c}{\textbf{CS}} \\
    Joint & $97.00_{\pm0.10}$ & $82.15_{\pm3.06}$ & $73.01_{\pm4.73}$ & \multicolumn{3}{c}{-} \\
    Bare model & $61.22_{\pm0.10}$ & $43.14_{\pm5.24}$ & $41.59_{\pm10.05}$ & $-45.55_{\pm0.02}$ & $-43.15_{\pm6.05}$ & $-31.36_{\pm11.32}$ \\
    LwF & $92.51_{\pm4.10}$ & $57.48_{\pm5.07}$ & $62.28_{\pm6.75}$ & $-5.49_{\pm4.78}$ & $-30.89_{\pm6.60}$ & $-4.27_{\pm7.36}$ \\
    ERGNN & $94.40_{\pm0.14}$ & $61.32_{\pm3.09}$ & $60.26_{\pm3.97}$ & $-3.26_{\pm0.29}$ & $-29.69_{\pm8.80}$ & $-15.50_{\pm2.88}$ \\
    DSLR & \textcolor{blue}{\underline{$96.47_{\pm0.80}$}} & \textcolor{blue}{\underline{$81.85_{\pm1.56}$}} &
    $65.78_{\pm2.70}$ &\textcolor{blue} {\underline{$-2.01_{\pm0.88}$}} & \textcolor{blue}{\underline{$-6.70_{\pm2.16}$}} & $-5.04_{\pm4.40}$ \\
    DMSG & $93.87_{\pm1.14}$ & $68.28_{\pm1.72}$ & \textcolor{blue}{\underline{$74.68_{\pm3.88}$}} & $-4.24_{\pm1.43}$ & $-15.91_{\pm4.15}$ & \textcolor{red}{$\mathbf{4.32_{\pm8.61}}$} \\
    SCE & $78.90_{\pm7.20}$ & $48.26_{\pm14.97}$ & $54.99_{\pm8.05}$ & $-22.48_{\pm8.69}$ & $-52.39_{\pm9.29}$ & $-27.48_{\pm17.28}$ \\
    CLNode & $93.84_{\pm0.32}$ & $49.72_{\pm1.92}$ & $48.17_{\pm1.45}$ & $-3.84_{\pm0.47}$ & $-57.87_{\pm2.50}$ & $-58.73_{\pm1.96}$ \\
    TSS & $79.62_{\pm5.16}$ & $48.28_{\pm4.03}$ & $49.41_{\pm12.24}$ & $-14.29_{\pm8.06}$ & $-47.73_{\pm2.55}$ & $-31.95_{\pm11.31}$ \\
    TFR & $94.11_{\pm1.98}$ & $39.92_{\pm10.13}$ & $42.09_{\pm3.92}$ & $-3.43_{\pm2.58}$ & $-50.30_{\pm11.79}$ & $-38.35_{\pm8.88}$ \\
    UFO & \textcolor{red}{$\mathbf{97.29_{\pm0.29}}$} & \textcolor{red}{$\mathbf{91.33_{\pm0.32}}$} & \textcolor{red}{$\mathbf{77.70_{\pm1.83}}$} & \textcolor{red}{$\mathbf{-1.30_{\pm0.27}}$} & \textcolor{red}{$\mathbf{0.81_{\pm0.31}}$} & \textcolor{blue}{\underline{$-0.84_{\pm1.79}$}} \\
    \bottomrule
  \end{tabular}
\end{table}

\begin{figure}[H]
\centering
\setlength{\tabcolsep}{4pt}
\footnotesize
\begin{tabular}{ccccc}
\includegraphics[width=0.18\linewidth]{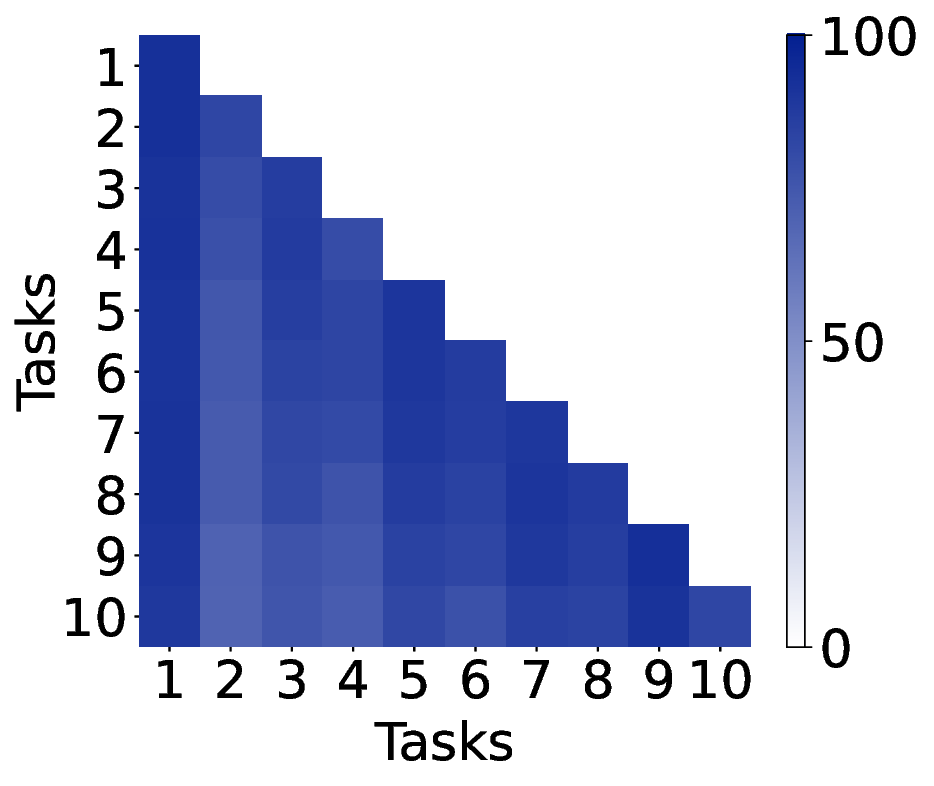} &
\includegraphics[width=0.18\linewidth]{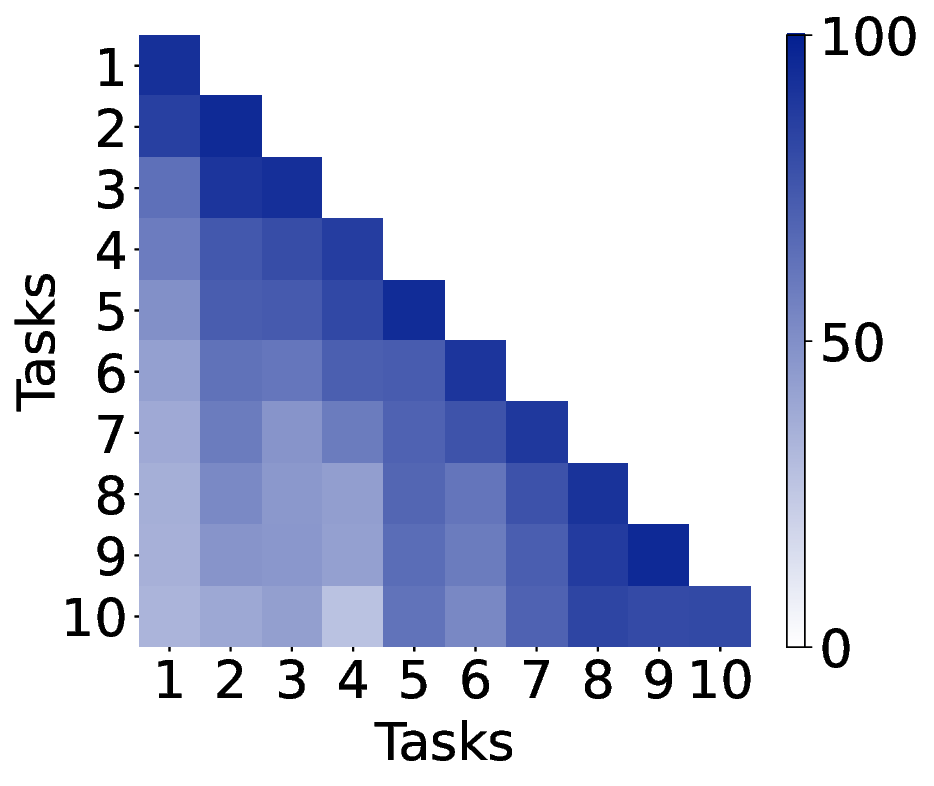} &
\includegraphics[width=0.18\linewidth]{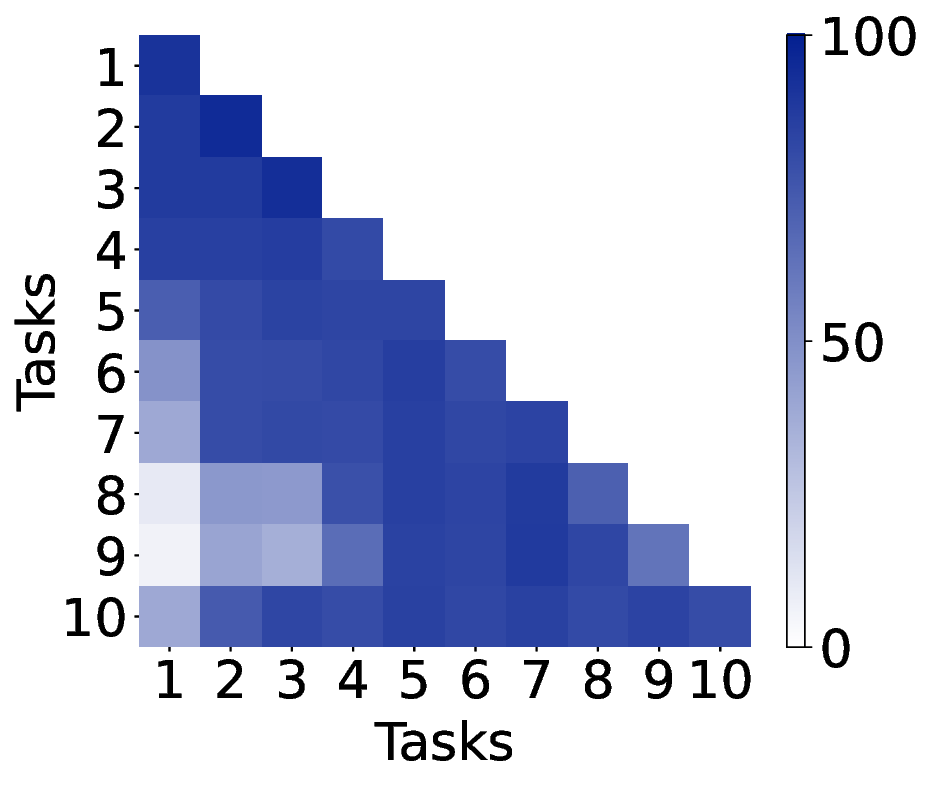} &
\includegraphics[width=0.18\linewidth]{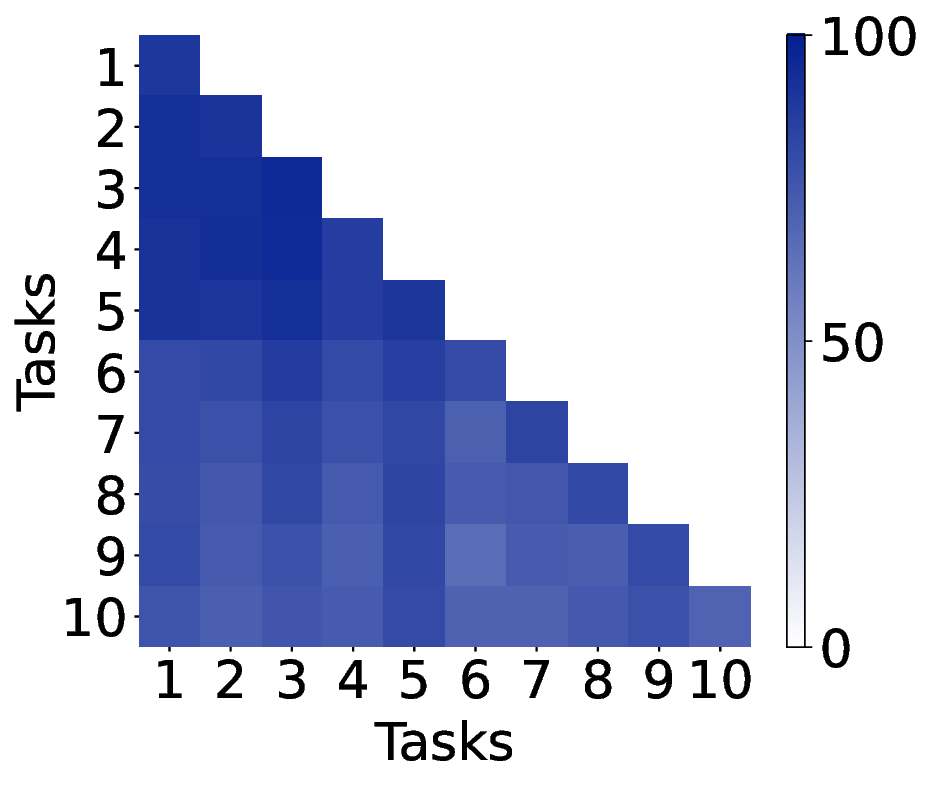} &
\includegraphics[width=0.18\linewidth]{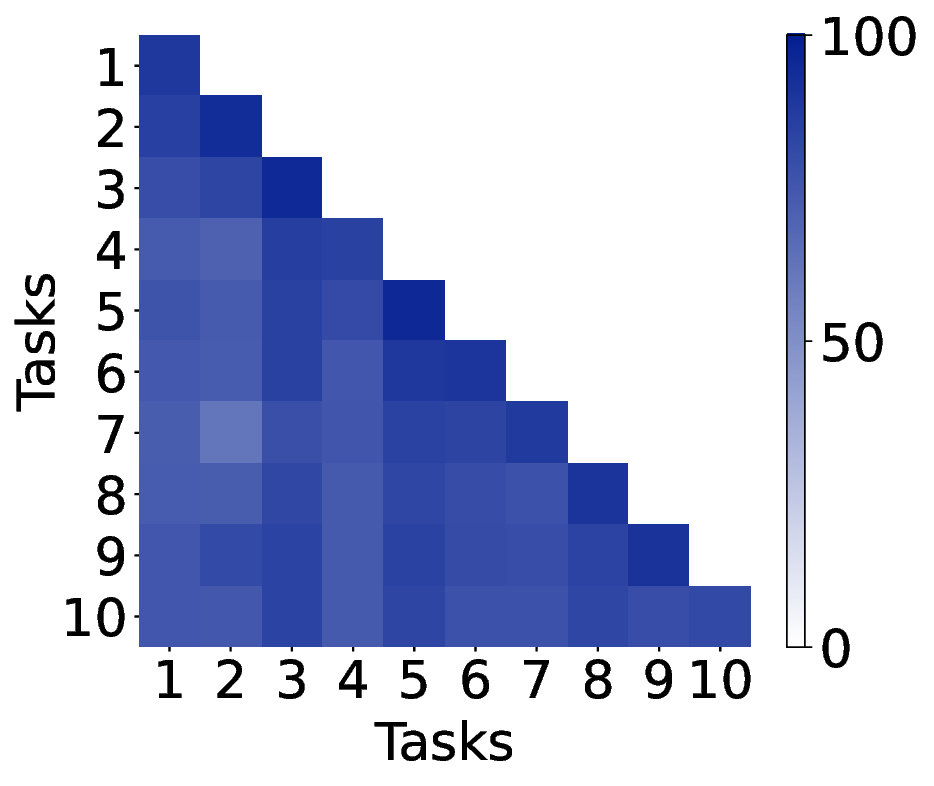} \\
UFO & LwF & ERGNN & DSLR & DMSG \\
\includegraphics[width=0.18\linewidth]{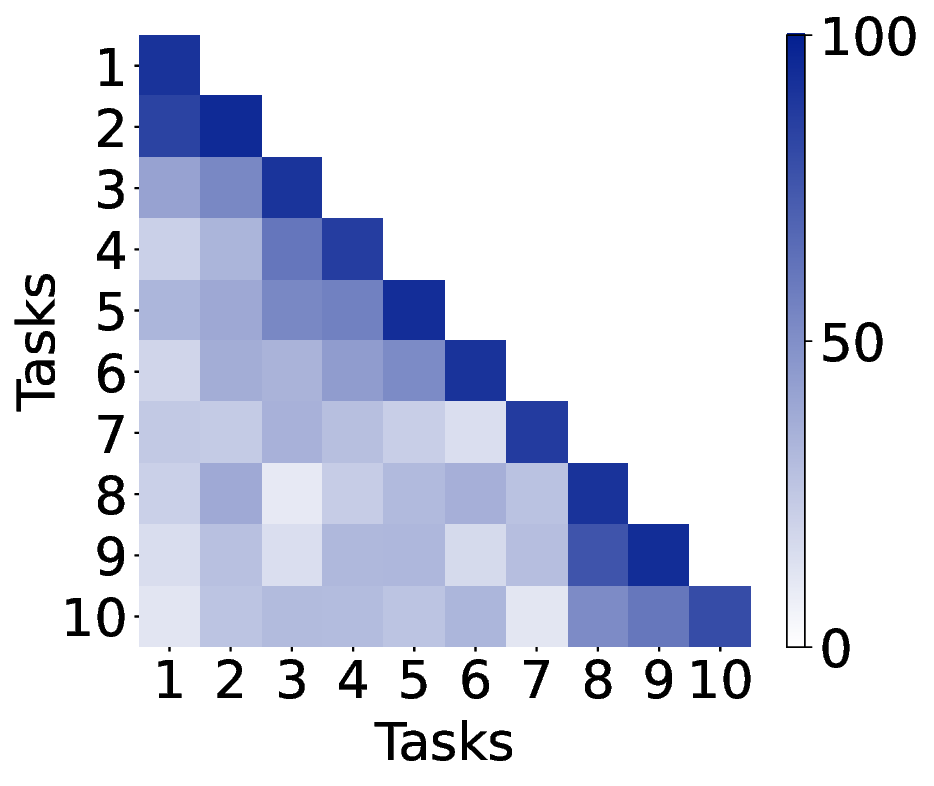} &
\includegraphics[width=0.18\linewidth]{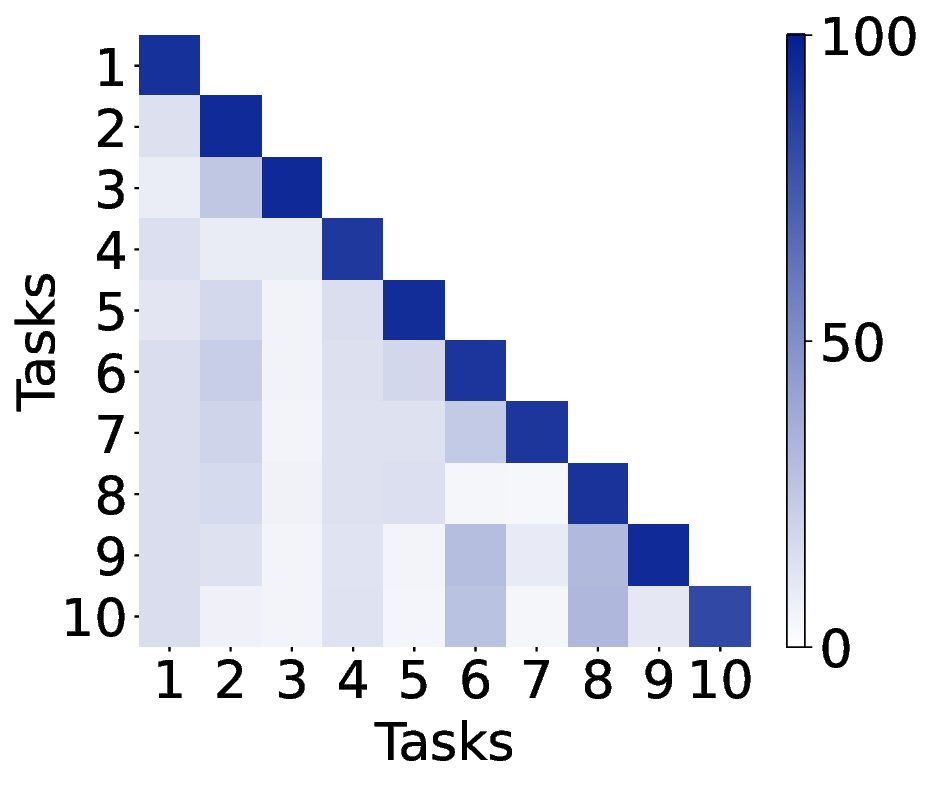} &
\includegraphics[width=0.18\linewidth]{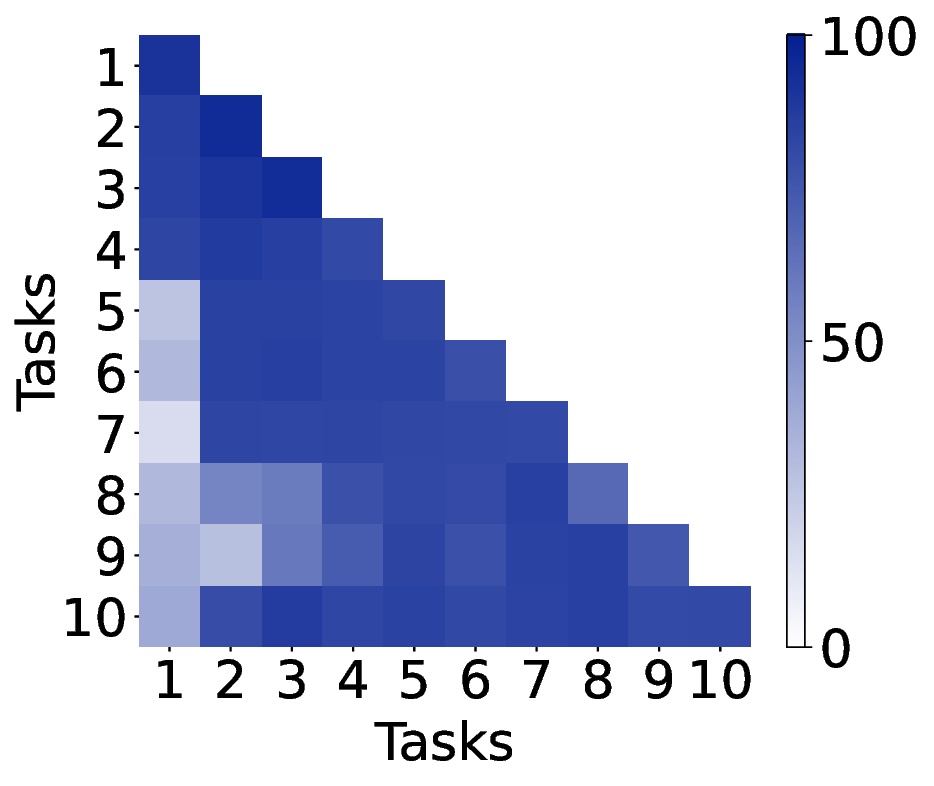} &
\includegraphics[width=0.18\linewidth]{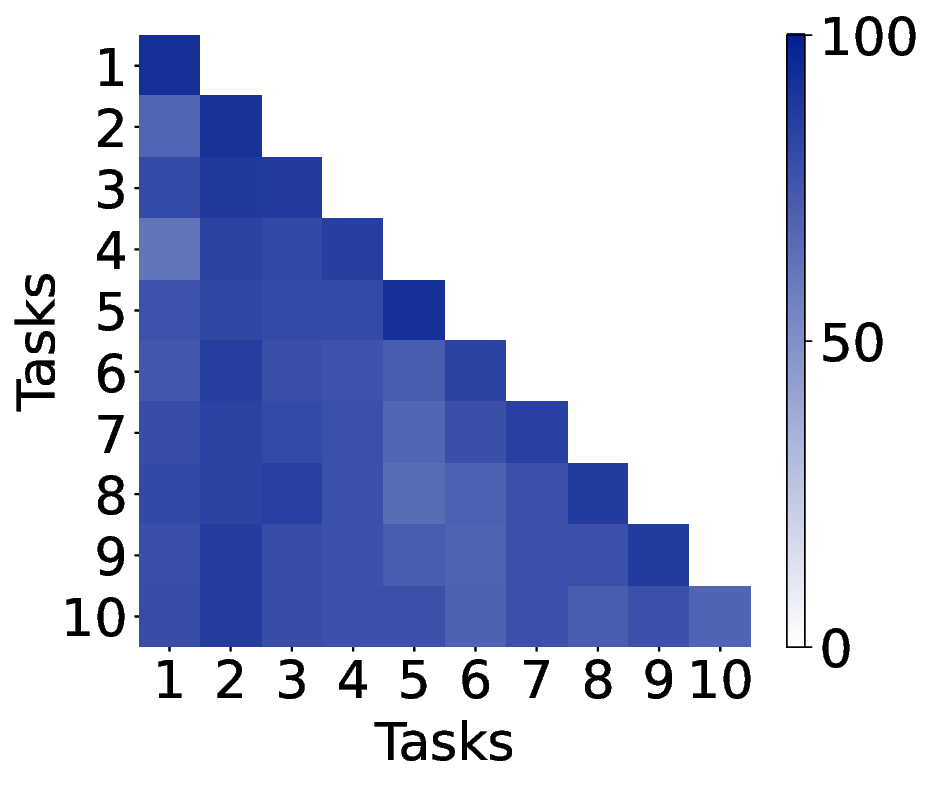} &
\includegraphics[width=0.18\linewidth]{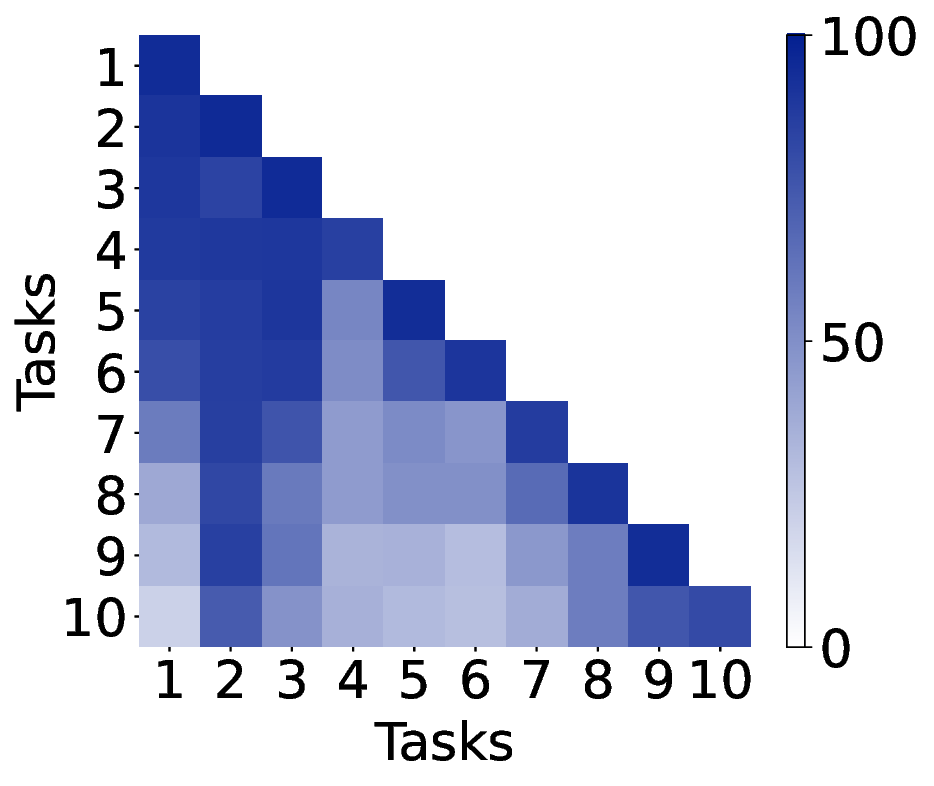} \\
Bare Model & SCE & CLNode & TSS & TFR
\end{tabular}
\caption{Performance matrices of various methods on CoraFull with $0$\% noise.}
\label{fig:methods_0}
\end{figure}

\begin{figure}[H]
\centering
\setlength{\tabcolsep}{4pt}
\footnotesize
\begin{tabular}{ccccc}
\includegraphics[width=0.18\linewidth]{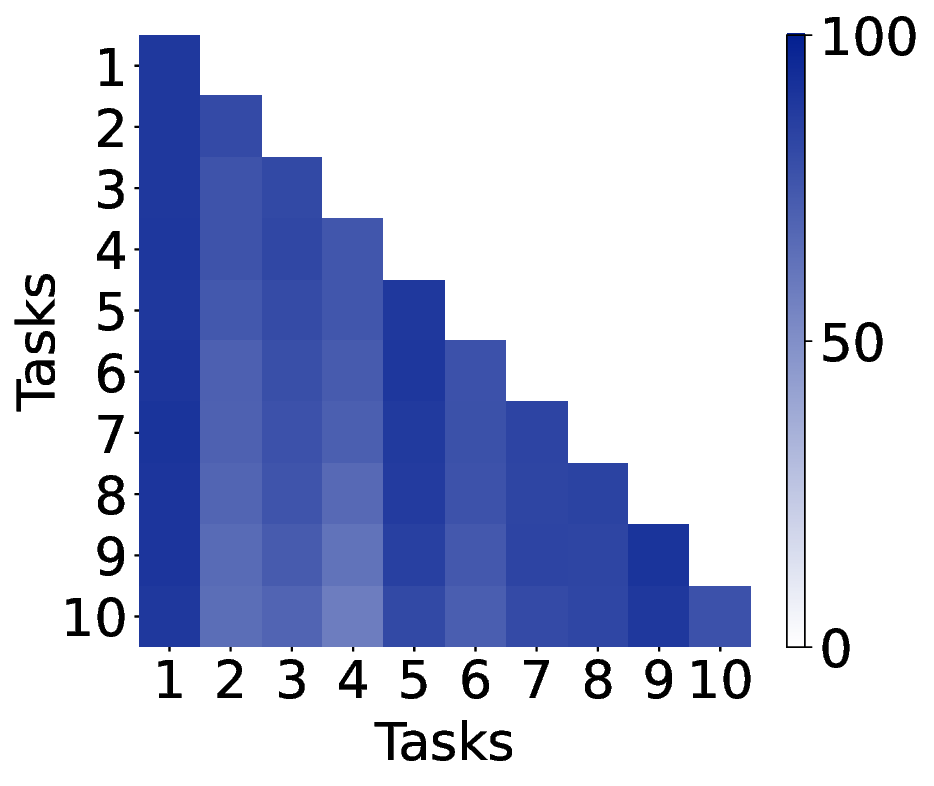} &
\includegraphics[width=0.18\linewidth]{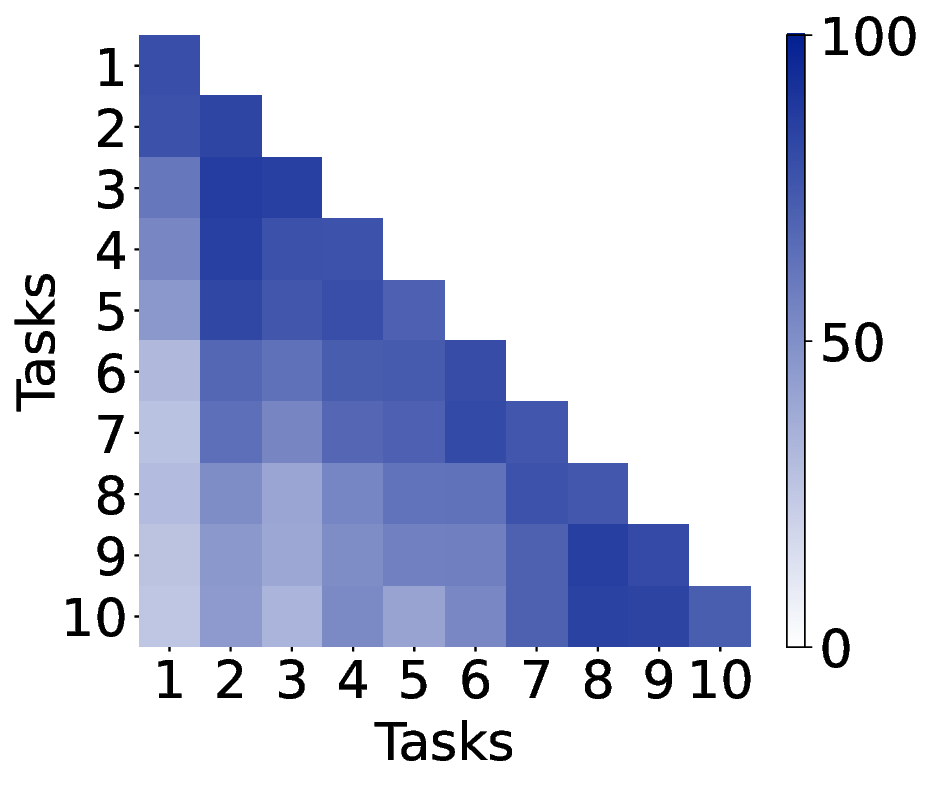} &
\includegraphics[width=0.18\linewidth]{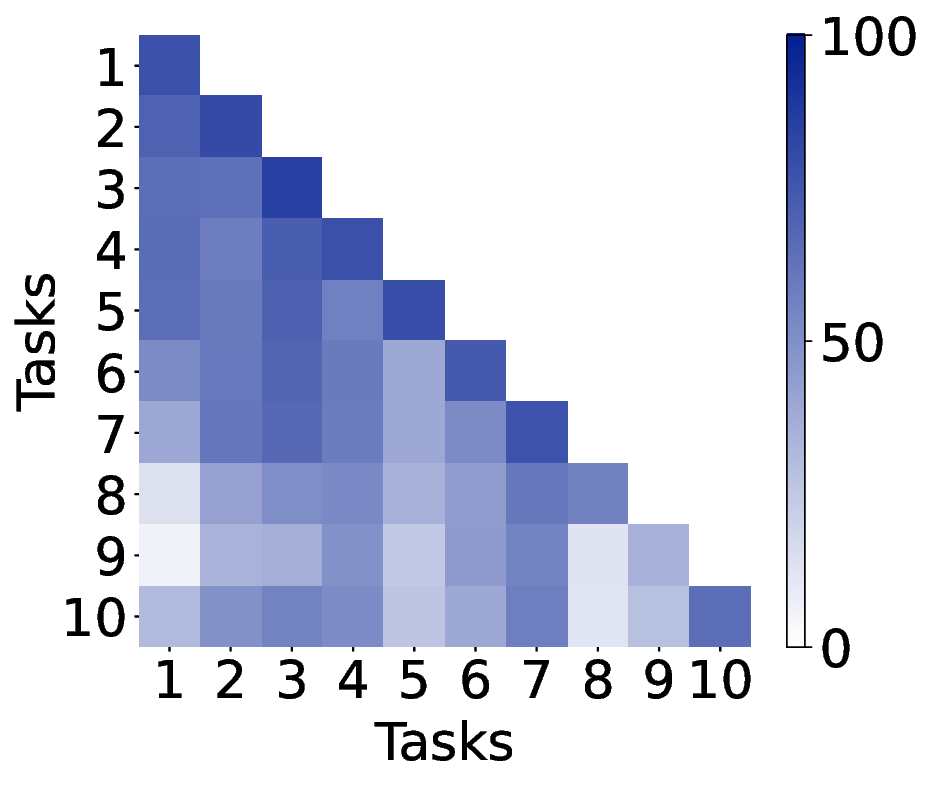} &
\includegraphics[width=0.18\linewidth]{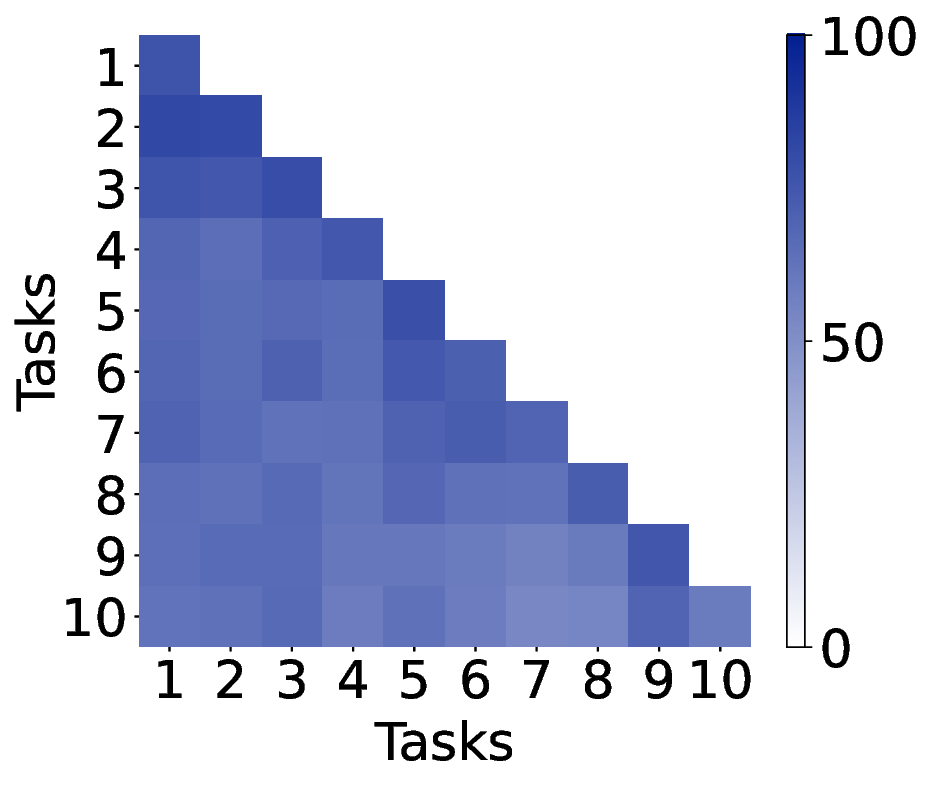} &
\includegraphics[width=0.18\linewidth]{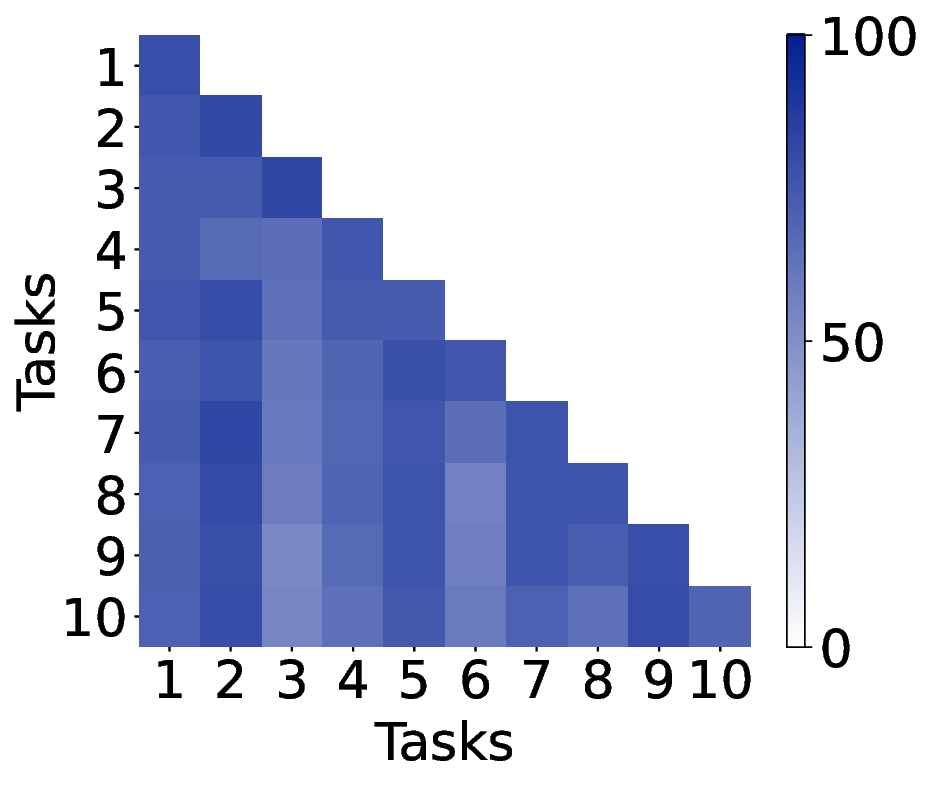} \\
UFO & LwF & ERGNN & DSLR & DMSG \\
\includegraphics[width=0.18\linewidth]{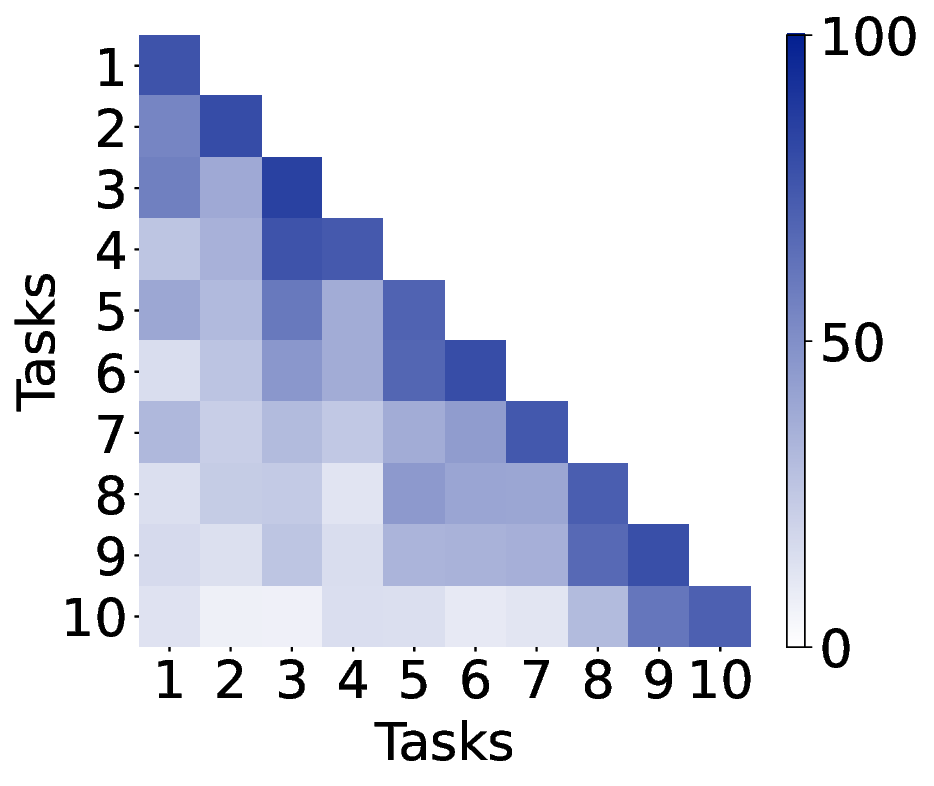} &
\includegraphics[width=0.18\linewidth]{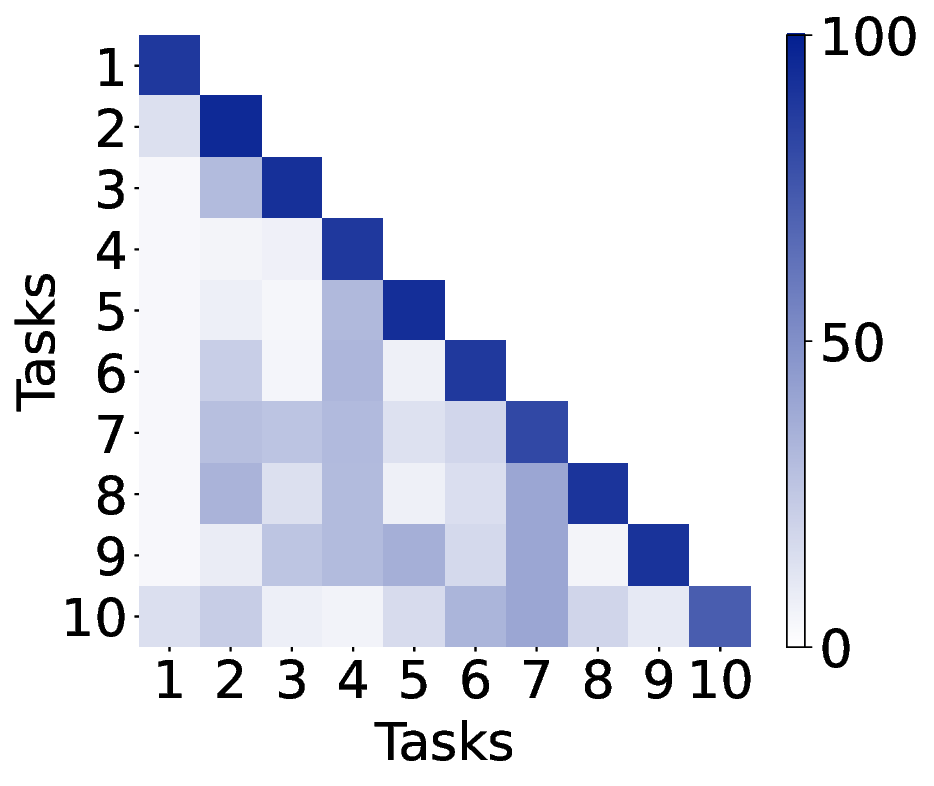} &
\includegraphics[width=0.18\linewidth]{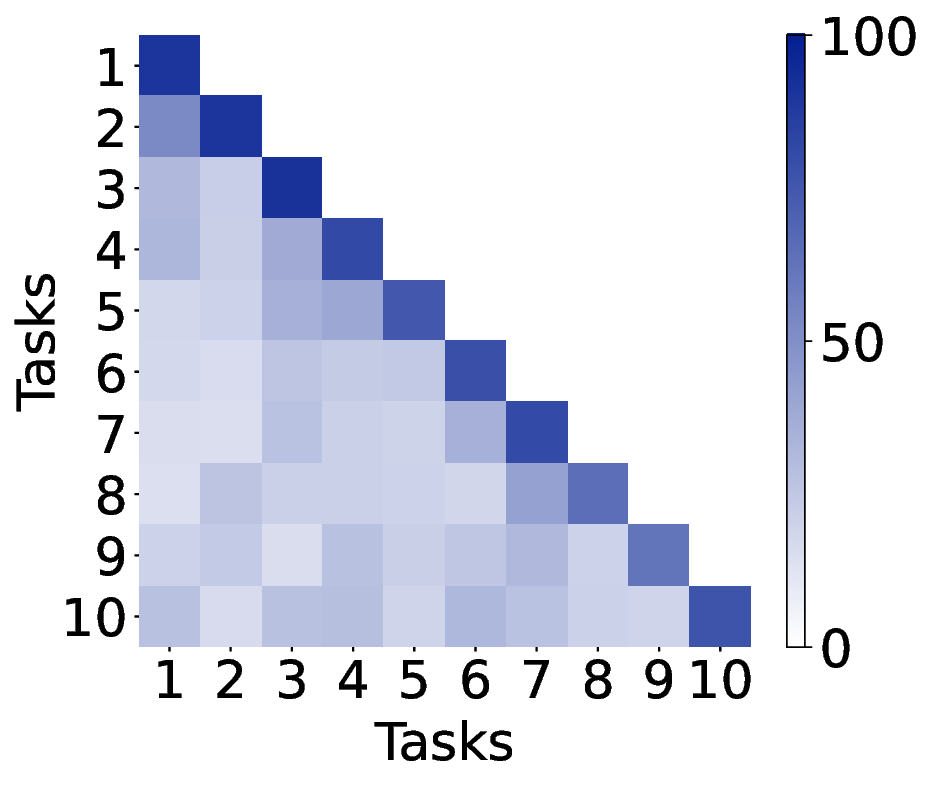} &
\includegraphics[width=0.18\linewidth]{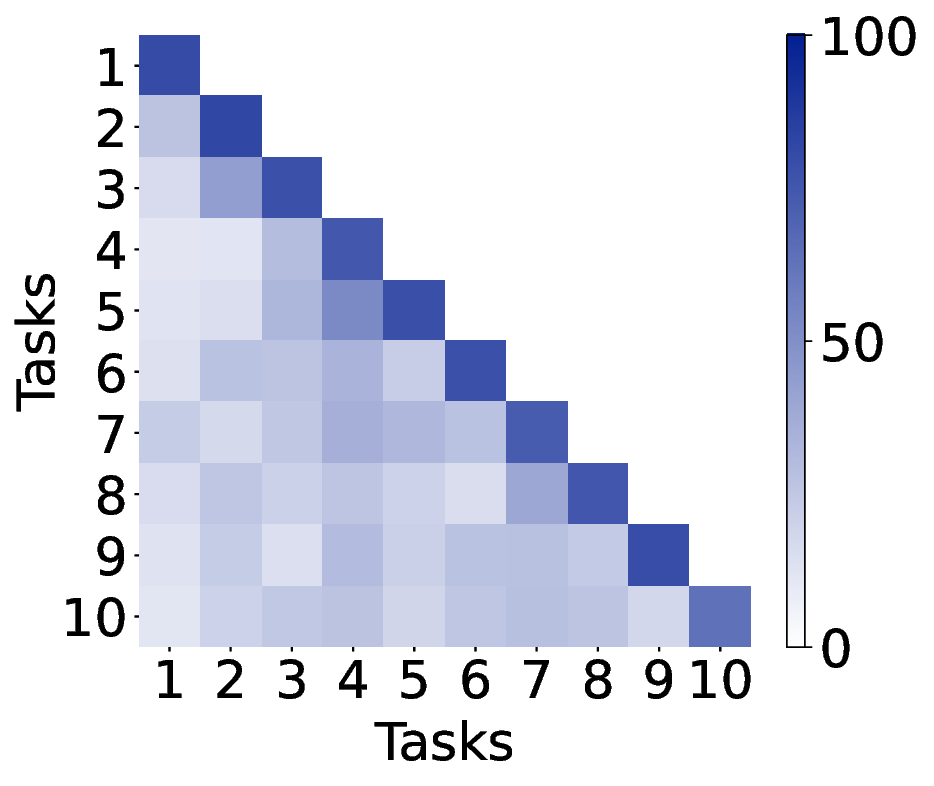} &
\includegraphics[width=0.18\linewidth]{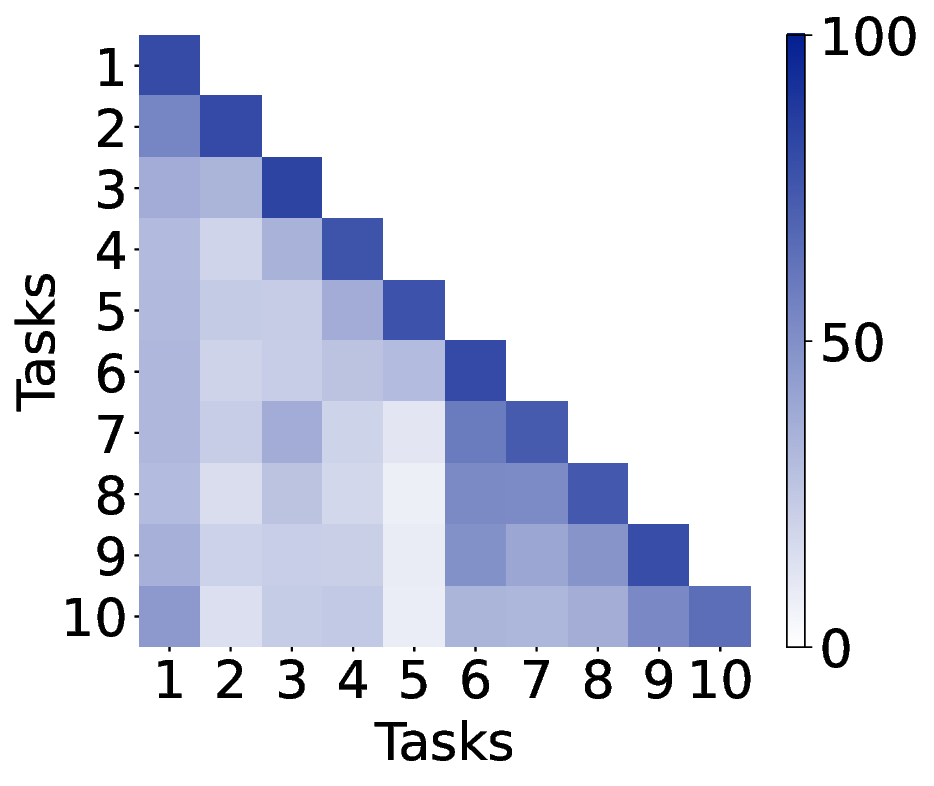} \\
Bare Model & SCE & CLNode & TSS & TFR
\end{tabular}
\caption{Performance matrices of various methods on CoraFull with 15\% noise.}
\label{fig:methods_15}
\end{figure}

\begin{figure}[H]
\centering
\footnotesize
\setlength{\tabcolsep}{4pt}
\begin{tabular}{ccccc}
     \includegraphics[width=0.18\linewidth]{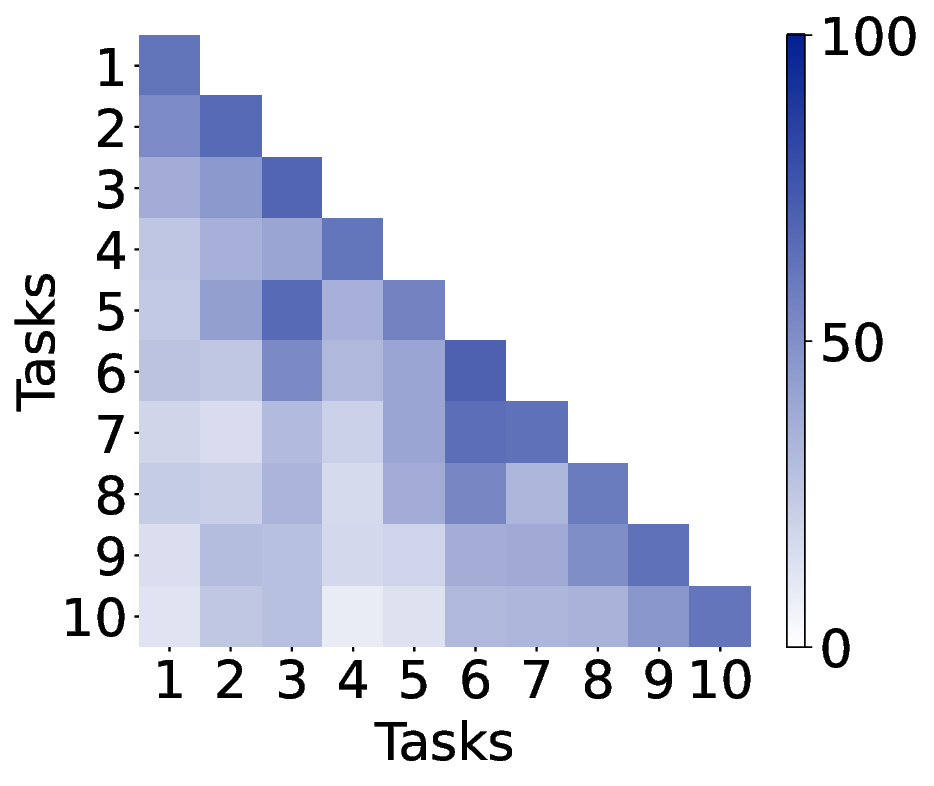} &
     \includegraphics[width=0.18\linewidth]{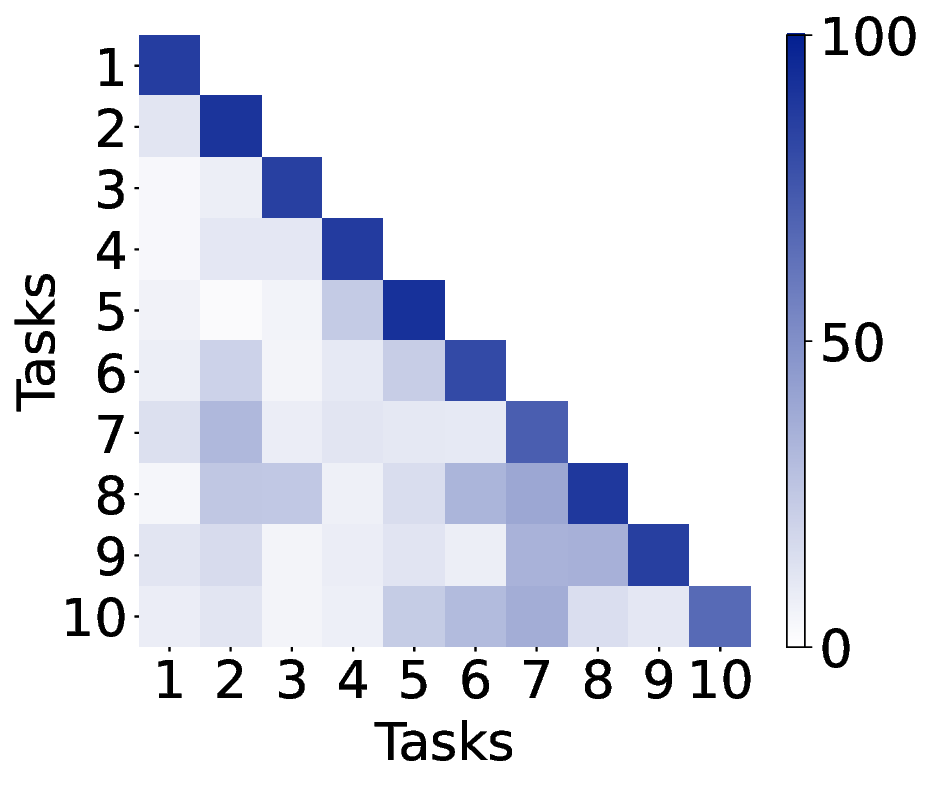} &
     \includegraphics[width=0.18\linewidth]{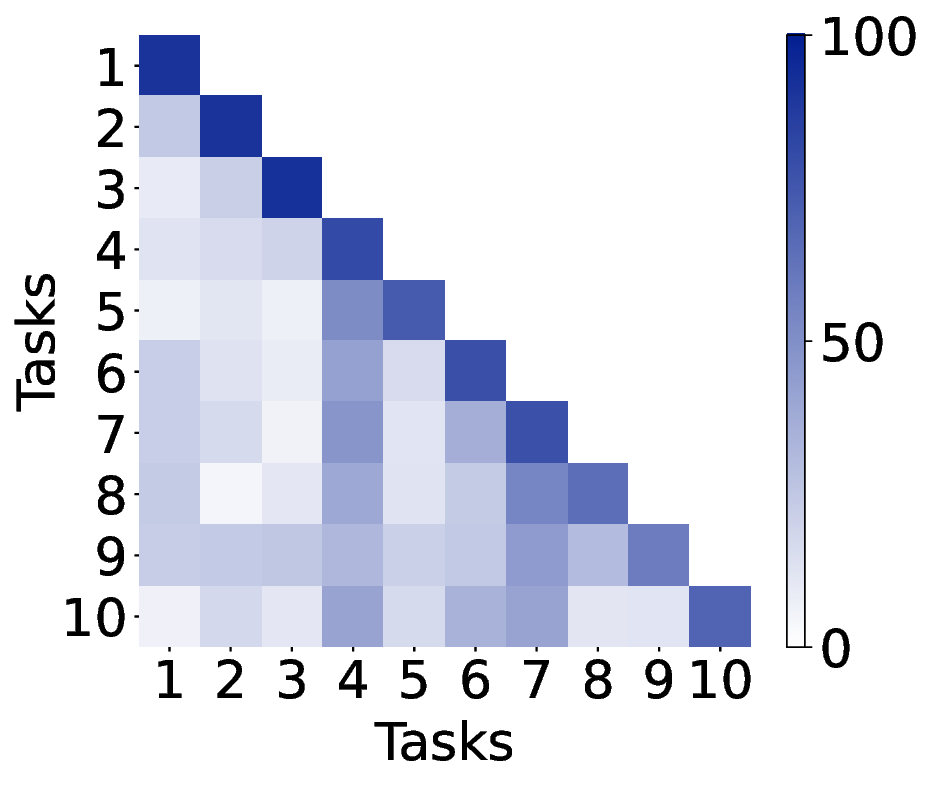} &
     \includegraphics[width=0.18\linewidth]{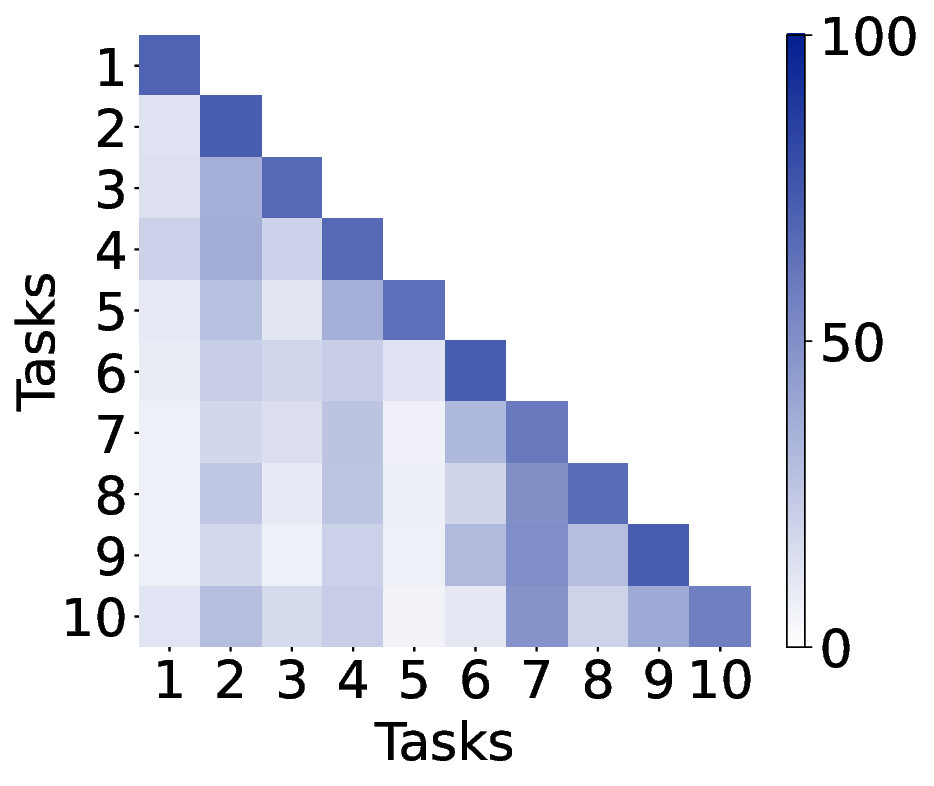} &
     \includegraphics[width=0.18\linewidth]{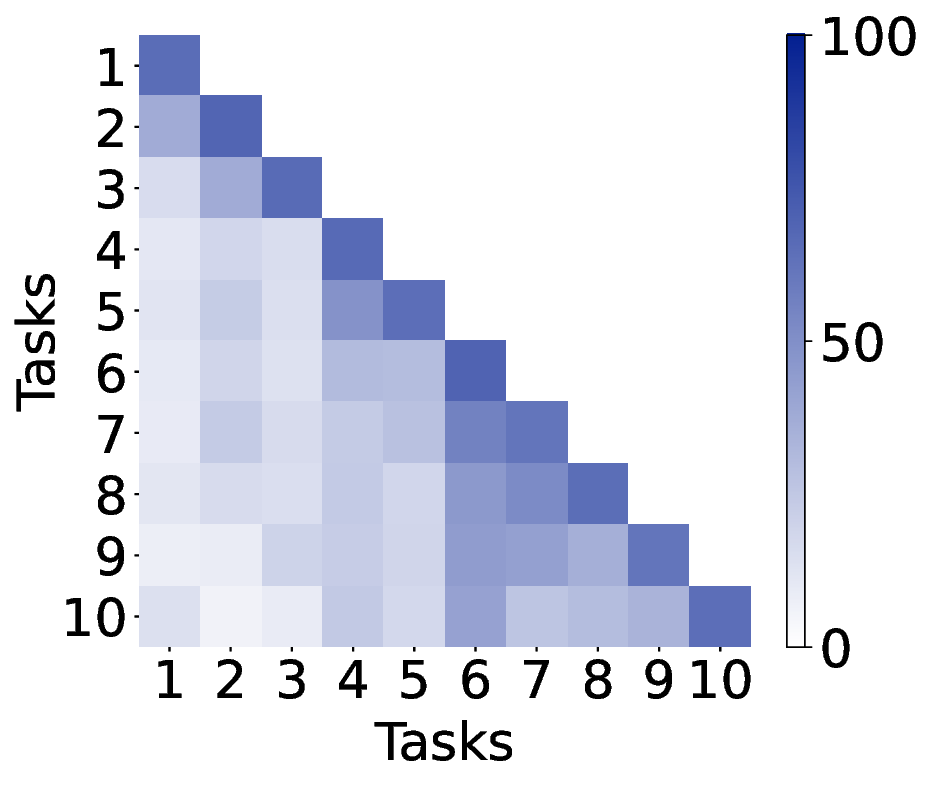} \\
     Bare Model & SCE & CLNode & TSS & TFR
\end{tabular}
\caption{Performance matrices of various methods on CoraFull with $30$\% noise.}
\label{fig:methods5}
\end{figure}

\begin{figure}[H]
    \centering
    \includegraphics[width=0.95\linewidth]{Figure/legend.png}
    \begin{subfigure}{\linewidth}
        \centering
        \includegraphics[width=0.95\linewidth]{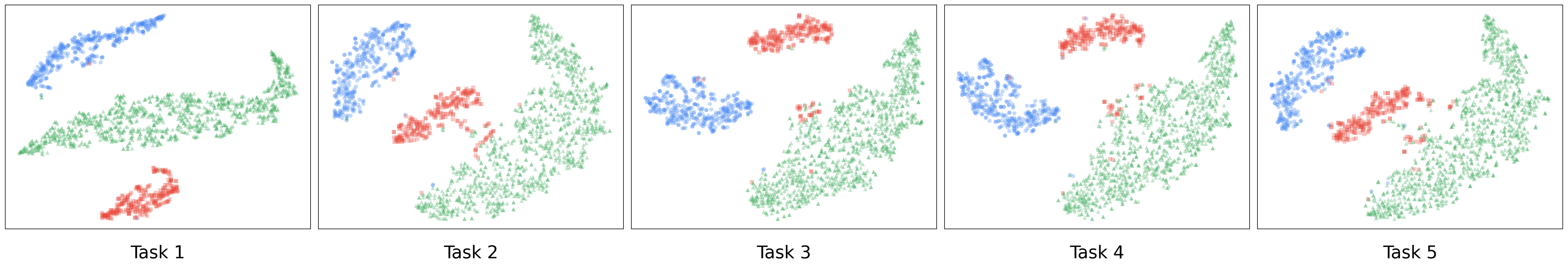}
        \subcaption{UFO with $0$\% noise}
    \end{subfigure}
    \begin{subfigure}{\linewidth}
        \centering
        \includegraphics[width=0.95\linewidth]{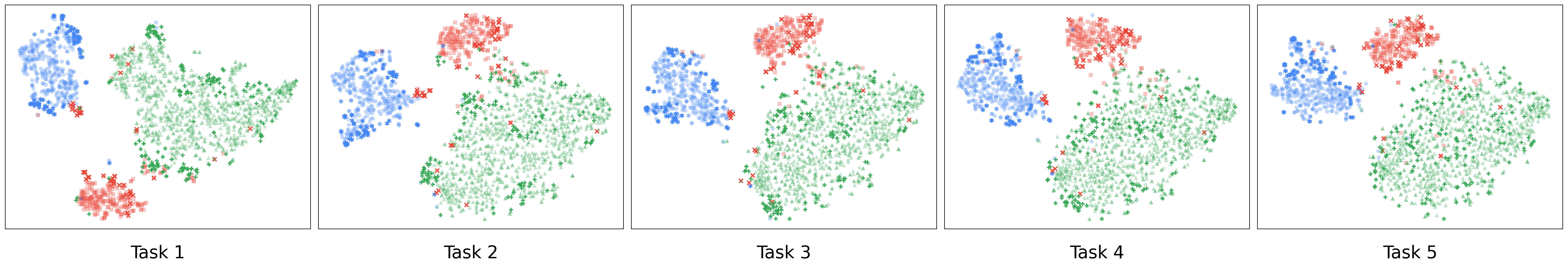}
        \subcaption{UFO with $15$\% noise}
    \end{subfigure}
    \begin{subfigure}{\linewidth}
        \centering
        \includegraphics[width=0.95\linewidth]{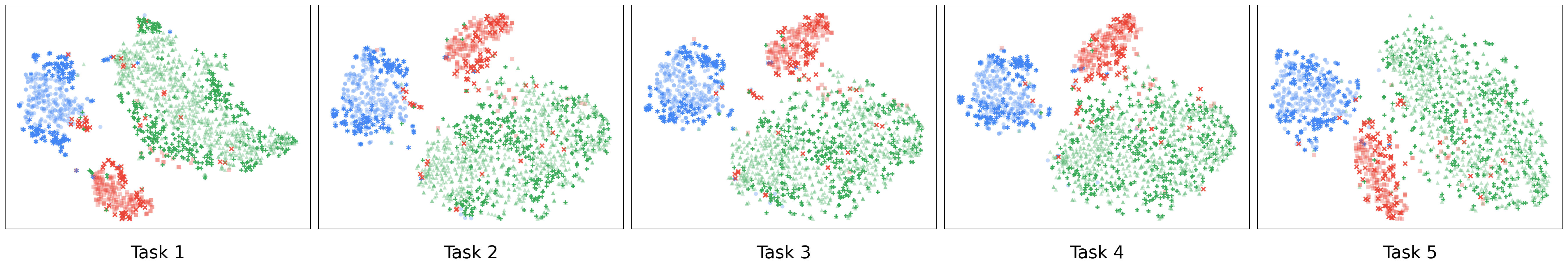}
        \subcaption{UFO with $30$\% noise}
    \end{subfigure}
    \caption{t-SNE visualization of the evolution of embeddings for classes $0$, $1$, and $2$ on sequential CS.}
    \label{fig:t_SNE222}
\end{figure}

\begin{figure}[H]
    \centering
    \includegraphics[width=0.95\linewidth]{Figure/legend.png}
    \begin{subfigure}{\linewidth}
        \centering
        \includegraphics[width=0.95\linewidth]{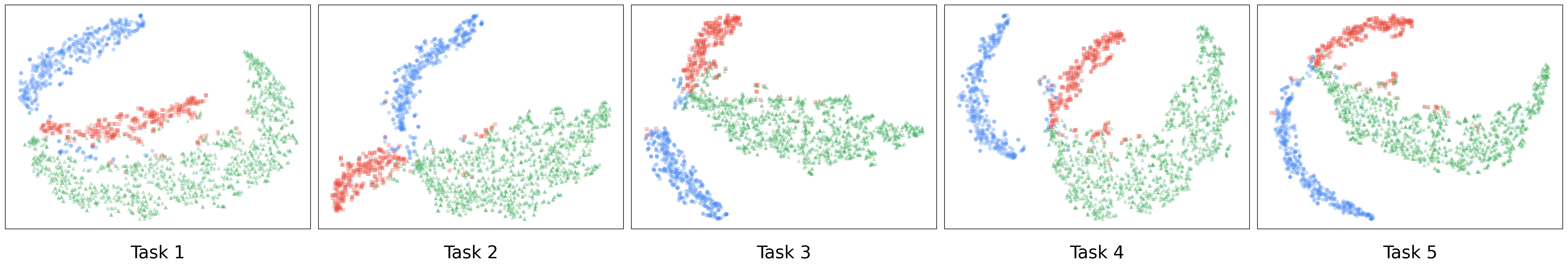}
        \caption{Joint with $0$\% noise}
    \end{subfigure}
    \begin{subfigure}{\linewidth}
        \centering
        \includegraphics[width=0.95\linewidth]{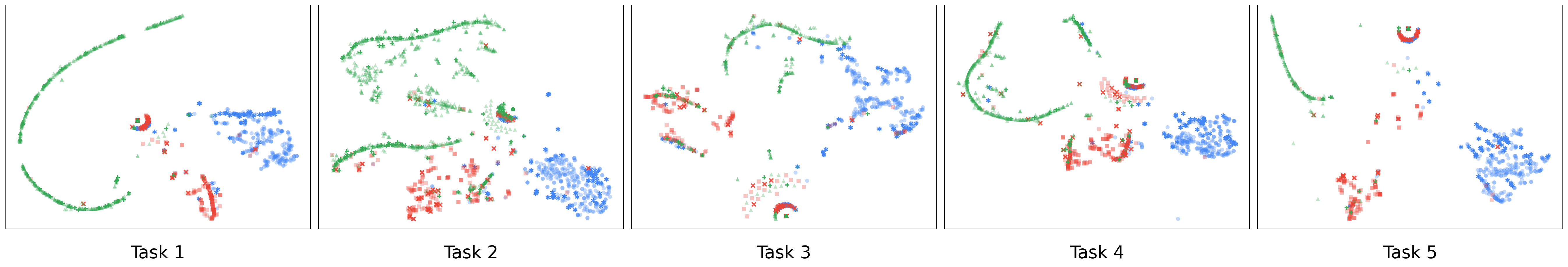}
        \caption{Joint with $15$\% noise}
    \end{subfigure}
    \begin{subfigure}{\linewidth}
        \centering
        \includegraphics[width=0.95\linewidth]{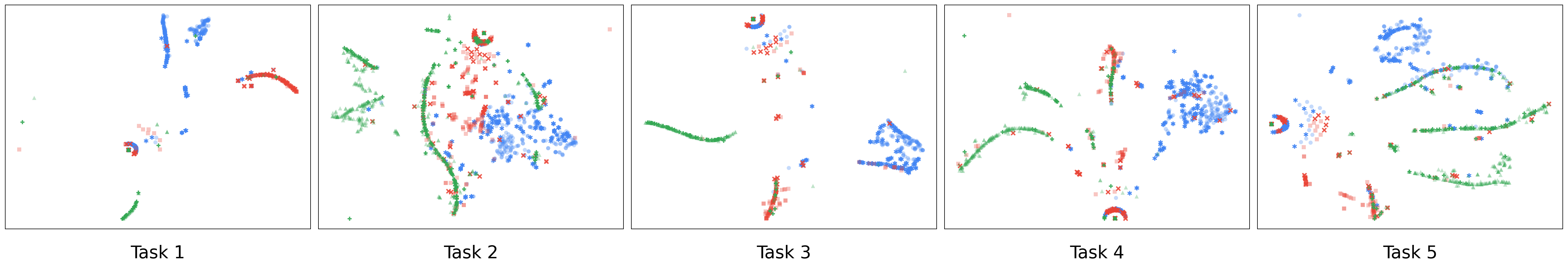}
        \caption{Joint with $30$\% noise}
    \end{subfigure}
    \caption{t-SNE visualization of the evolution of embeddings for classes 0, 1, and 2 on sequential CS.}
    \label{fig:t_SNE}
\end{figure}

\section{Additional Experimental Results}
\subsection{Pair Flipping Noise} \label{app:pair}
In many scenarios, noisy labels are not randomly assigned to any class, but are more likely to be confused with very similar classes.
For example, in the CS dataset \cite{shchur2018pitfalls}, authors in paired research fields, such as machine learning and data mining or robotics and computer vision, may share similar research topics.
To simulate this, we further conduct experiments under pair flipping noise with ratios of $0$\%, $15$\%, and $30$\%.
The results are reported in Table~\ref{tab:dataset_subscript_pm}, and it can be observed that UFO remains effective in preserving performance under structured label corruption.

\subsection{Performance Matrices}
\label{app:visualization}

Figures~\ref{fig:methods_0}--\ref{fig:methods5} present additional performance matrices on CoraFull under different noise ratios.
As can be seen, most baselines show clear degradation on earlier tasks as the task sequence progresses, and this degradation becomes more evident when the noise ratio increases.
Although some methods, such as SCE, CLNode, TSS, and TFR, produce darker diagonal regions, the lighter lower-left regions indicate limited preservation of previous task knowledge under noisy supervision.
This suggests that improving robustness to noisy labels alone is not sufficient for robust continual graph learning,
which further supports the need for a unified framework such as UFO to handle knowledge preservation and noisy supervision.

\subsection{t-SNE Visualizations}
\label{ref:t-sne-joint}
Figures~\ref{fig:t_SNE222} and~\ref{fig:t_SNE} show the complete visualizations of UFO and Joint on sequential CS.
As can be seen, UFO maintains more compact and stable class clusters as the task sequence progresses, even when the noise ratio increases.
Noisy nodes are mainly distributed near the cluster boundaries or outer regions, indicating that UFO can reduce their disturbance to the main class structure.
In contrast, Joint shows more scattered embeddings under noisy supervision, especially at higher noise ratios, where the class structures become less clear.
These visualizations further demonstrate the effectiveness of UFO.

% \section{Technical appendices and supplementary material}
% Technical appendices with additional results, figures, graphs, and proofs may be submitted with the paper submission before the full submission deadline (see above). You can upload a ZIP file for videos or code, but do not upload a separate PDF file for the appendix. There is no page limit for the technical appendices. 

% Note: Think of the appendix as ``optional reading'' for reviewers. The paper must be able to stand alone without the appendix; for example, adding critical experiments that support the main claims to an appendix is inappropriate. 

%%%%%%%%%%%%%%%%%%%%%%%%%%%%%%%%%%%%%%%%%%%%%%%%%%%%%%%%%%%%

% \newpage
% \input{checklist.tex}

\end{document}